\documentclass[10pt,journal,compsoc]{IEEEtran}
\ifCLASSOPTIONcompsoc
  \usepackage[nocompress]{cite}
\else
  \usepackage{cite}
\fi
\usepackage{hyperref} 
\hypersetup{
    colorlinks=true,
    linkcolor=red,
}
\usepackage{amssymb}
\usepackage{bbding}

\ifCLASSINFOpdf
   \usepackage[pdftex]{graphicx}
\else
\fi

\usepackage{stfloats}
\usepackage{xcolor}
\usepackage{color}

\usepackage{multirow}

\usepackage{caption}
\usepackage{subcaption}

\begin{document}
%
\title{Unsupervised Point Cloud Representation Learning with Deep Neural Networks: A Survey}

\author{Aoran~Xiao$^*$,
        Jiaxing~Huang$^*$,
        Dayan~Guan,
        Xiaoqin~Zhang,
        Shijian~Lu,
        and~Ling~Shao~\IEEEmembership{Fellow,~IEEE}
\IEEEcompsocitemizethanks{
\IEEEcompsocthanksitem Aoran~Xiao and Jiaxing~Huang are co-first authors.
\IEEEcompsocthanksitem Aoran~Xiao, Jiaxing~Huang and Shijian~Lu are with the School of Computer Science and Engineering, Nanyang Technological University, Singapore. 
\IEEEcompsocthanksitem Dayan~Guan is with Mohamed bin Zayed University of Artificial Intelligence, United Arab Emirates.
\IEEEcompsocthanksitem Xiaoqin Zhang is with Key Laboratory of Intelligent Informatics for Safety \& Emergency of Zhejiang Province, Wenzhou University, China.
\IEEEcompsocthanksitem Ling Shao is with UCAS-Terminus AI Lab, UCAS.
\IEEEcompsocthanksitem Corresponding authors: Shijian~Lu (shijian.lu@ntu.edu.sg) and Xiaoqin Zhang (zhangxiaoqinnan@gmail.com)
}
}

%
%

\markboth{IEEE Transactions on Pattern Analysis and Machine Intelligence}%
{Shell \MakeLowercase{\textit{et al.}}: Bare Demo of IEEEtran.cls for Computer Society Journals}
%



\IEEEtitleabstractindextext{%
\begin{abstract}
Point cloud data have been widely explored due to its superior accuracy and robustness under various adverse situations. Meanwhile, deep neural networks (DNNs) have achieved very impressive success in various applications such as surveillance and autonomous driving. The convergence of point cloud and DNNs has led to many deep point cloud models, largely trained under the supervision of large-scale and densely-labelled point cloud data. 
Unsupervised point cloud representation learning, which aims to learn general and useful point cloud representations from unlabelled point cloud data, has recently attracted increasing attention due to the constraint in large-scale point cloud labelling. This paper provides a comprehensive review of unsupervised point cloud representation learning using DNNs. It first describes the motivation, general pipelines as well as terminologies of the recent studies. Relevant background including widely adopted point cloud datasets and DNN architectures is then briefly presented. This is followed by an extensive discussion of existing unsupervised point cloud representation learning methods according to their technical approaches. We also quantitatively benchmark and discuss the reviewed methods over multiple widely adopted point cloud datasets. Finally, we share our humble opinion about several challenges and problems that could be pursued in the future research in unsupervised point cloud representation learning. A project associated with this survey has been built at \url{https://github.com/xiaoaoran/3d_url_survey}. 
\end{abstract}

}

\maketitle

\IEEEdisplaynontitleabstractindextext

%
\IEEEpeerreviewmaketitle

\IEEEraisesectionheading{\section{Introduction}\label{sec:introduction}}
3D acquisition technologies have experienced fast development in recent years. This can be witnessed by different 3D sensors that have become increasingly popular in both industrial and our daily lives such as LiDAR sensors in autonomous vehicles, RGB-D cameras in Kinect and Apple devices, 3D scanners in various reconstruction tasks, etc. Meanwhile, 3D data of different modalities such as meshes, point clouds, depth images and volumetric grids, which capture accurate geometric information for both objects and scenes, have been collected and widely applied in different areas such as autonomous driving, robotics, medical treatment, remote sensing, etc.

\begin{figure}[t]
    \centering
    \includegraphics[width=0.5\textwidth]{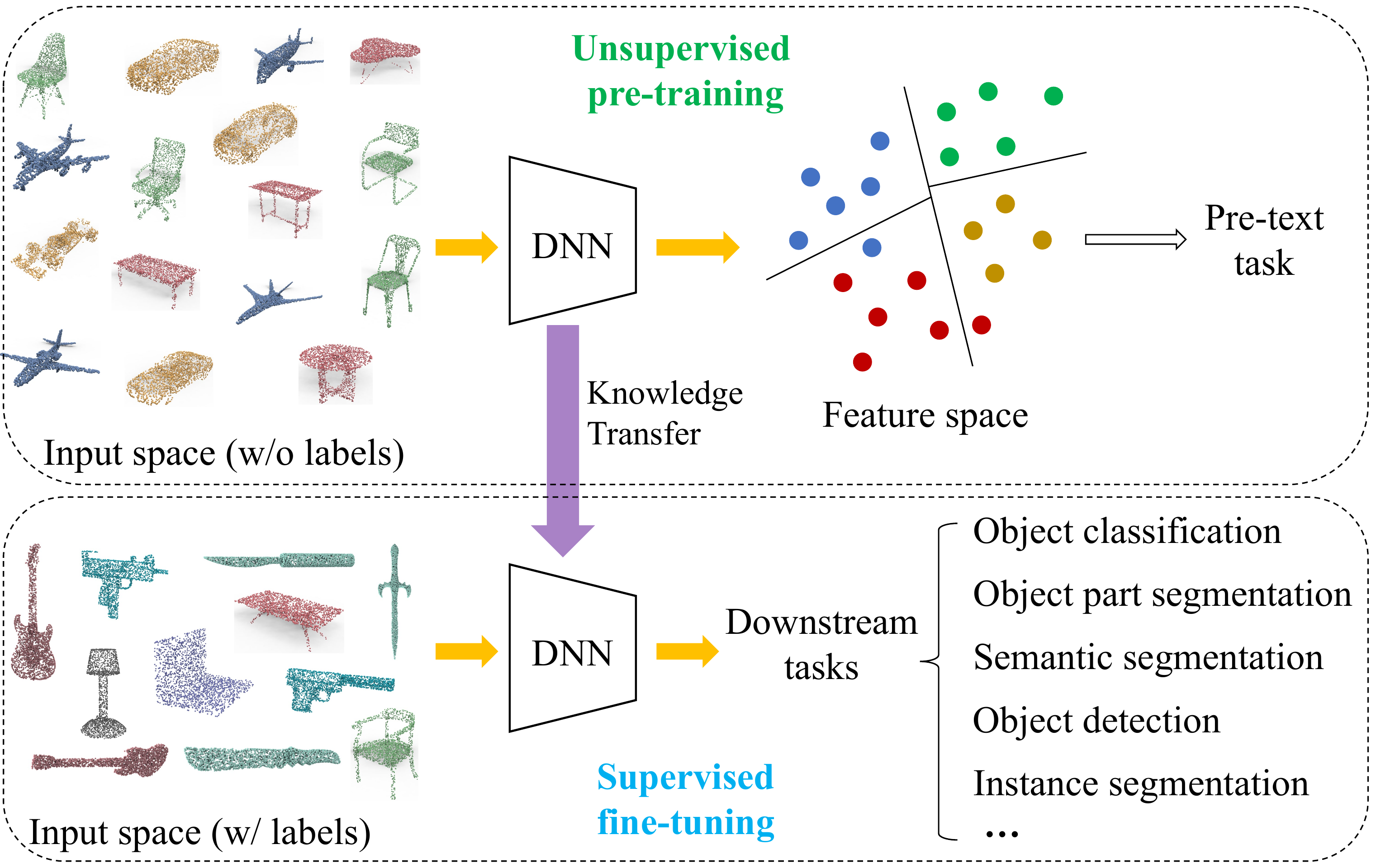}
    \caption{The general pipeline of unsupervised representation learning on point clouds: Deep neural networks are first pre-trained with unannotated point clouds via unsupervised learning over certain pre-text tasks. The learned unsupervised point cloud representations are then transferred to various downstream tasks to provide network initialization, with which the pre-trained networks are \textit{fine-tuned} with a small amount of annotated task-specific point cloud data.}
    \label{fig.PC-URL}
\end{figure}

Point cloud as one source of ubiquitous and widely used 3D data can be directly captured with entry-level depth sensors before triangulating into meshes or converting to voxels. This makes it easily applicable to various 3D scene understanding tasks~\cite{huang2021spatio} such as 3D object detection and shape analysis, semantic segmentation, etc. With the advance of deep neural networks (DNNs), point cloud understanding has attracted increasing attention as observed by a large number of deep architectures and deep models developed in recent years~\cite{guo2020deep}. On the other hand, effective training of deep networks requires large-scale human-annotated training data such as 3D bounding boxes for object detection and point-wise annotations for semantic segmentation, which are usually laborious and time-consuming to collect due to 3D view changes and visual inconsistency between human perception and point cloud display. Efficient collection of large-scale annotated point clouds has become one bottleneck for effective design, evaluations, and deployment of deep networks while handling various real-world tasks~\cite{hou2021exploring}.

\begin{figure*}[t]
    \centering
    \includegraphics[width=\textwidth]{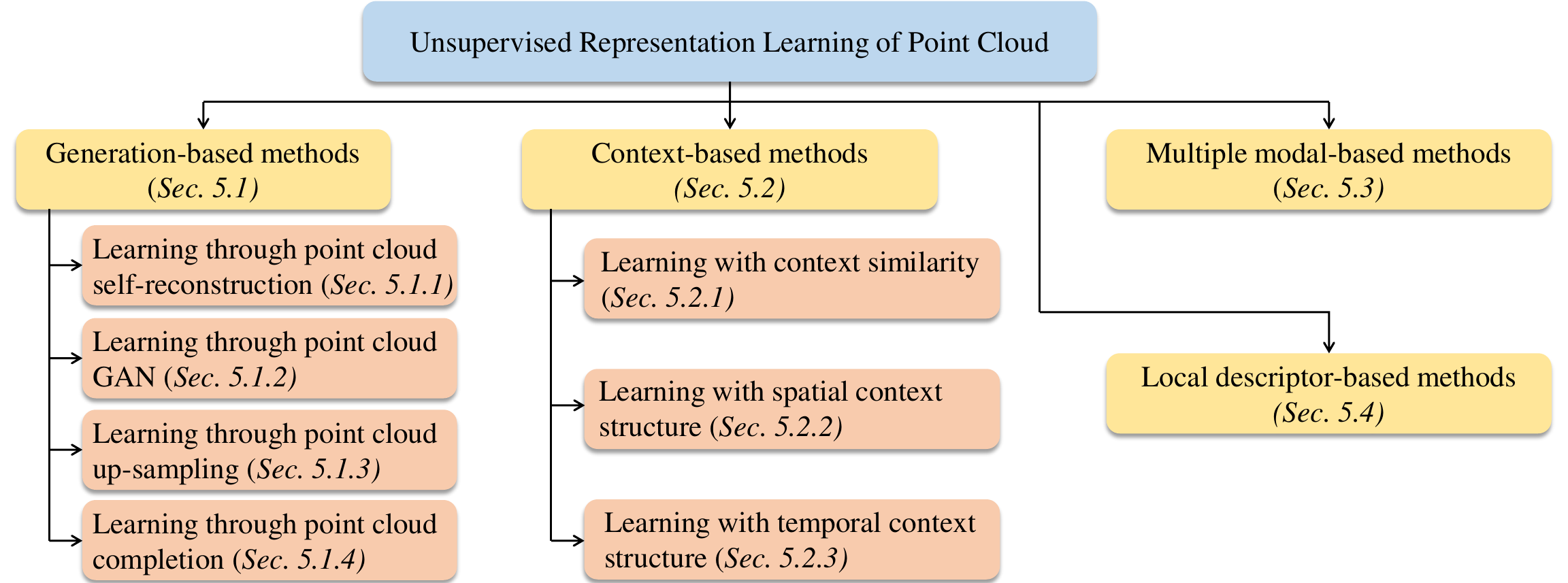}
    \caption{Taxonomy of existing unsupervised methods in point cloud representation learning.}
    \label{fig.landscape_PC-URL}
\end{figure*}

Unsupervised representation learning (URL), which aims to learn robust and general feature representations from unlabelled data, has recently been studied intensively for mitigating the laborious and time-consuming data annotation challenge. As Fig.~\ref{fig.PC-URL} shows, URL works in a similar way to pre-training which learns useful knowledge from unlabelled data and transfers the learned knowledge to various downstream tasks~\cite{jing2021self}. More specifically, URL can provide useful network initialization with which well-performing network models can be trained with a small amount of labelled and task-specific training data without suffering from much over-fitting as compared with training from random initialization. URL can thus help reduce training data and annotations which has demonstrated great effectiveness in the areas of natural language processing (NLP)~\cite{radford2019language,kenton2019bert}, 2D computer vision~\cite{he2020momentum,grill2020bootstrap,chen2020improved,he2022masked}, etc.

Similar to URL from other types of data such as texts and 2D images, URL of point clouds has recently attracted increasing attention in the computer vision research community. A number of URL techniques have been reported which are typically achieved by designing different pre-text tasks such as 3D object reconstruction~\cite{valsesia2018learning}, partial object completion~\cite{Wang_2021_ICCV}, 3D jigsaws solving~\cite{sauder2019self}, etc. However, URL of point clouds still lags far behind as compared with its counterparts in NLP and 2D computer vision tasks. For the time being, training from scratch on various target new data is still the prevalent approach in most existing 3D scene understanding development. At the other end, URL from point cloud data is facing increasing problems and challenges, largely due to the lack of large-scale and high-quality point cloud data, unified deep backbone architectures, generalizable technical approaches, as well as comprehensive public benchmarks.

In addition, URL for point clouds is still short of systematic survey that can offer a clear big picture about this new yet challenging task. To fill up this gap, this paper presents a comprehensive survey on the recent progress in unsupervised point cloud representation learning from the perspective of datasets, network architectures, technical approaches, performance benchmarking, and future research directions. As shown in Fig. \ref{fig.landscape_PC-URL}, we broadly group existing methods into four categories based on their pretext tasks, including URL methods using data generation, global and local contexts, multimodality data and local descriptors, more details to be discussed in the ensuing subsections.

The major contributions of this work are threefold:
\begin{enumerate}
    \item It presents a comprehensive review of the recent development in unsupervised point cloud representation learning. To the best of our knowledge, it is the \textit{first} survey that provides an overview and big picture for this exciting research topic. 
    \item It studies the most recent progress of unsupervised point cloud representation learning, including a comprehensive benchmarking and discussion of existing methods over multiple public datasets.
    \item It shares several research challenges and potential research directions that could be pursued in unsupervised point cloud representation learning.
\end{enumerate}

The rest of this survey is organized as follows: In Section \ref{Background}, we introduce background knowledge of unsupervised point cloud learning including term definition, common tasks of point cloud understanding and relevant surveys to this work.
Section \ref{sec.PC_dataset} introduces widely-used datasets and their characteristics. Section \ref{Sec.DNN} introduces commonly used deep point cloud architectures with typical models that are frequently used for point cloud URL. In Section \ref{Sec.URL} we systematically review the methods for point cloud URL. Section \ref{sec.Benchmarks} summarizes and compares the performances of existing methods on multiple benchmark datasets. At last, we list several promising future directions for unsupervised point cloud representation learning in Section \ref{Sec.Future}.

\section{Background}\label{Background}

\subsection{Basic concepts}
We first define all relevant terms and concepts that are to be used in the ensuing sections.

\noindent\textbf{Point cloud data:} A point cloud $P$ is a set of vectors $P=\{p_1,...,p_N\}$ where each vector represents one point $p_i=[C_i, A_i]$. Here, $C_i\in \mathbf{R}^{1\times 3}$ refers to 3D coordinate $(x_i,y_i,z_i)$ of the point, and $A_i$ refers to feature attributes of the point such as RGB values, LiDAR intensity, normal values, etc., which are optional and variational depending on 3D sensors as well as applications.

\noindent \textbf{Supervised learning:} Under the paradigm of deep learning, supervised learning aims to train deep network models by using labelled training data.

\noindent \textbf{Unsupervised learning:} 
Unsupervised learning aims to train networks by using unlabelled training data.

\noindent \textbf{Unsupervised representation learning:} URL is a subset of unsupervised learning. It aims to learn meaningful representations from data without using any data labels/annotations, where the learned representations can be transferred to different downstream tasks. Some literature alternatively uses the term ``self-supervised learning".

\noindent \textbf{Semi-supervised learning:} In semi-supervised learning, deep networks are trained with a small amount of labelled data and a large amount of unlabelled data. It aims to mitigate data annotation constraints by learning from a small amount of labelled data and a large amount of unlabelled data that have similar distributions.

\noindent \textbf{Pre-training:} 
Network pre-training learns with certain pre-text tasks over other datasets. The learned parameters are often employed for model initialization for further fine-tuning with various task-specific data.

\noindent \textbf{Transfer learning:} Transfer learning aims to transfer knowledge across tasks, modalities or datasets. A typical scenario related to this survey is to perform unsupervised learning for pre-training for transferring the learned knowledge from unlabelled data to various downstream networks.

\subsection{Common 3D understanding tasks}\label{para.eval}

This subsection introduces common 3D understanding tasks including \textit{object-level tasks} in object classification and object part segmentation and \textit{scene-level tasks} in 3D object detection, semantic segmentation and instance segmentation. These tasks have been widely adopted to evaluate the quality of point cloud representations that are learned via various unsupervised learning methods, which will be discussed in detail in Section~\ref{sec.Benchmarks}.

\begin{figure}[b]
    \centering
    \includegraphics[width=0.45\textwidth]{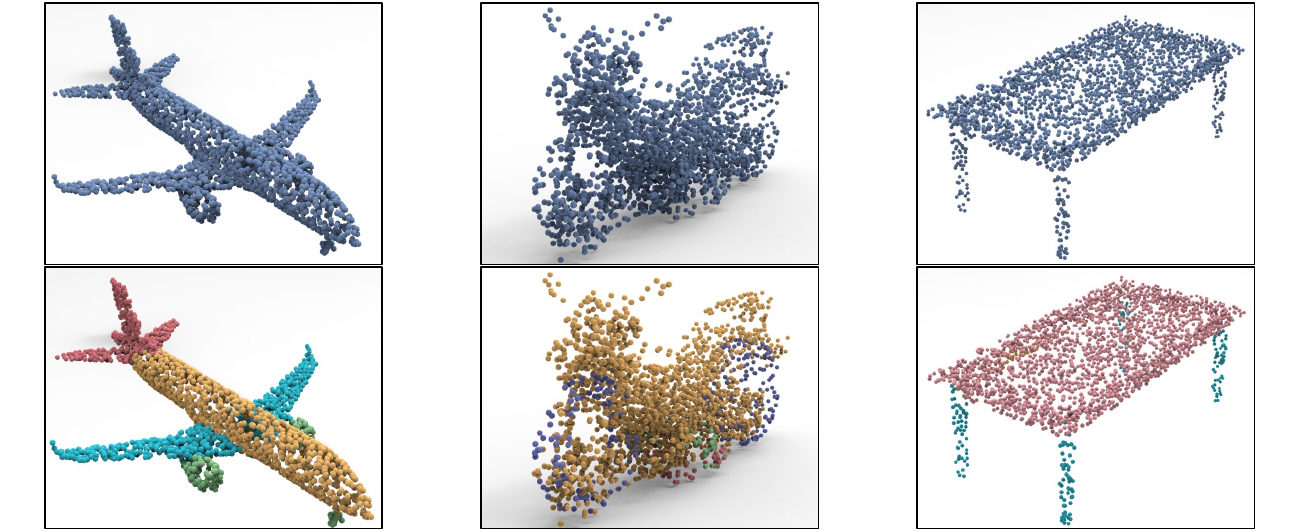}
    \caption{Illustration of object part segmentation: The first row shows a few object samples including \textit{airplane}, \textit{motorcycle}, and \textit{table} from the ShapeNetPart dataset \cite{chang2015shapenet}. The second row shows segmentation ground truth with different parts as highlighted by different colors.}
    \label{fig.part segmentation}
\end{figure}

\subsubsection{Object classification}\label{Sec.classification}
Object classification aims to classify point cloud objects into a number of pre-defined categories. Two evaluation metrics are most frequently used: The \textit{overall Accuracy} (OA) represents the averaged accuracy for all instances in the test set; The \textit{mean class accuracy} (mAcc) represents the mean accuracy of all object classes for the test set.

\subsubsection{Object part segmentation}\label{eval.partseg}

Object part segmentation is an important task for point cloud representation learning. It aims to assign a part category label (e.g., airplane wing, table leg, etc.) to each point as illustrated in Fig. \ref{fig.part segmentation}. The mean Intersection over Union (mIoU) \cite{qi2017pointnet} is the most widely adopted evaluation metric. For each instance, IoU is computed for each part belonging to that object category.  The mean of the part IoUs represents the IoU of that object instance.  The overall IoU is computed as the average of IoUs over all test instances while category-wise IoU (or class IoU) is calculated as the mean over instances under that category. 

\subsubsection{3D object detection}

\begin{figure}[h]
    \centering
    \begin{subfigure}[b]{0.2\textwidth}
        \includegraphics[width=\textwidth]{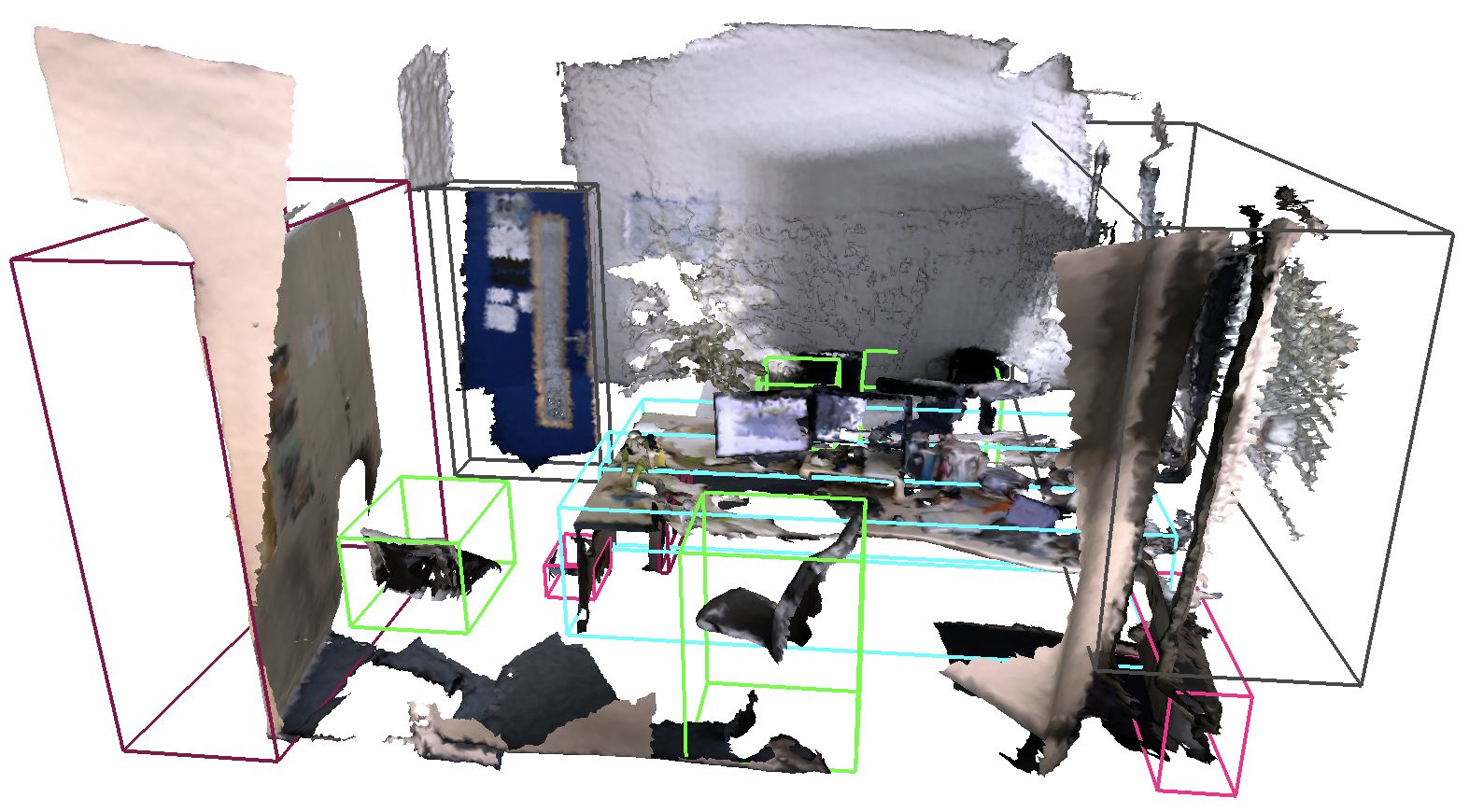}
        \caption{ScanNet-V2 dataset}
    \end{subfigure}
    \hspace{5mm}
    \begin{subfigure}[b]{0.2\textwidth}
        \includegraphics[width=\textwidth]{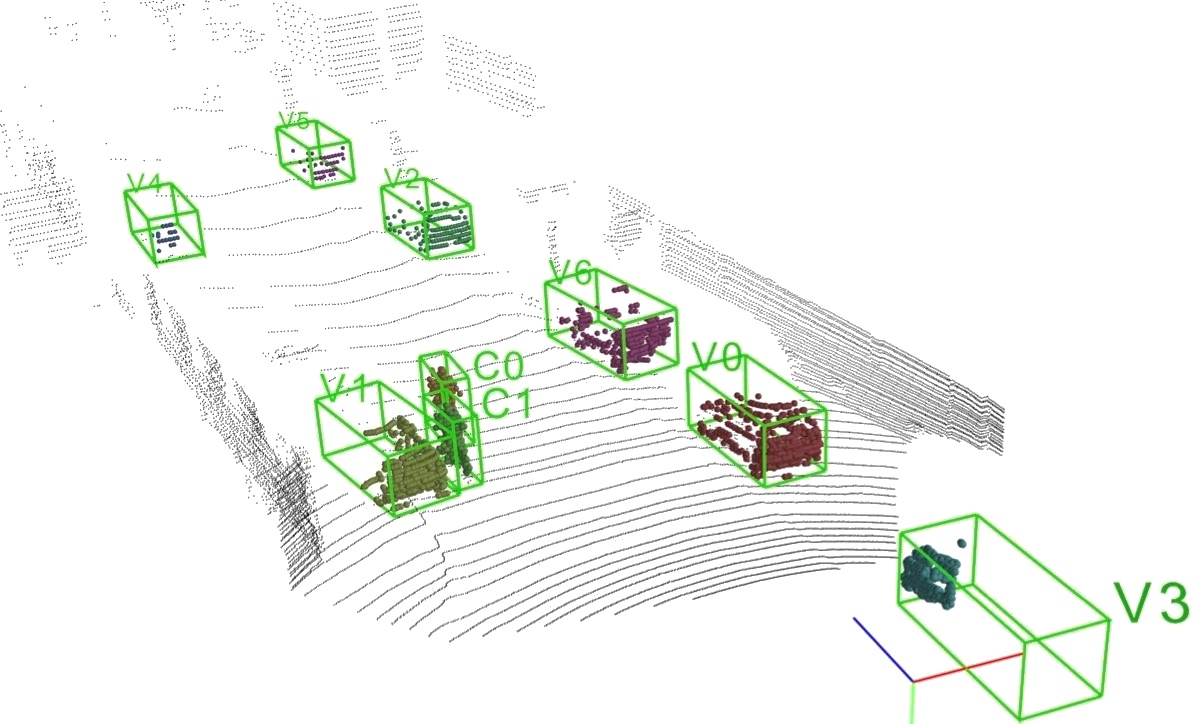}
        \caption{KITTI dataset}
    \end{subfigure}
    \caption{Illustration of 3D bounding boxes in point cloud object detection: The two graphs show 3D bounding boxes in datasets ScanNet-V2~\cite{dai2017scannet} and KITTI~\cite{geiger2013vision} which are cropped from \cite{qi2019votenet} and \cite{qi2018frustum}, respectively.}
    \label{fig.detection}
\end{figure}

3D object detection on point clouds is a crucial and indispensable task for many real-world applications, such as autonomous driving and domestic robots. The task aims to localize objects in the 3D space, \textit{i.e.} 3D object bounding boxes as illustrated in Fig. \ref{fig.detection}. The average precision (AP) metric has been widely used for evaluations in 3D object detection \cite{qi2019votenet,shi2019pointrcnn}.

\begin{figure}[b]
    \centering
    \begin{subfigure}[b]{0.2\textwidth}
        \includegraphics[width=\textwidth]{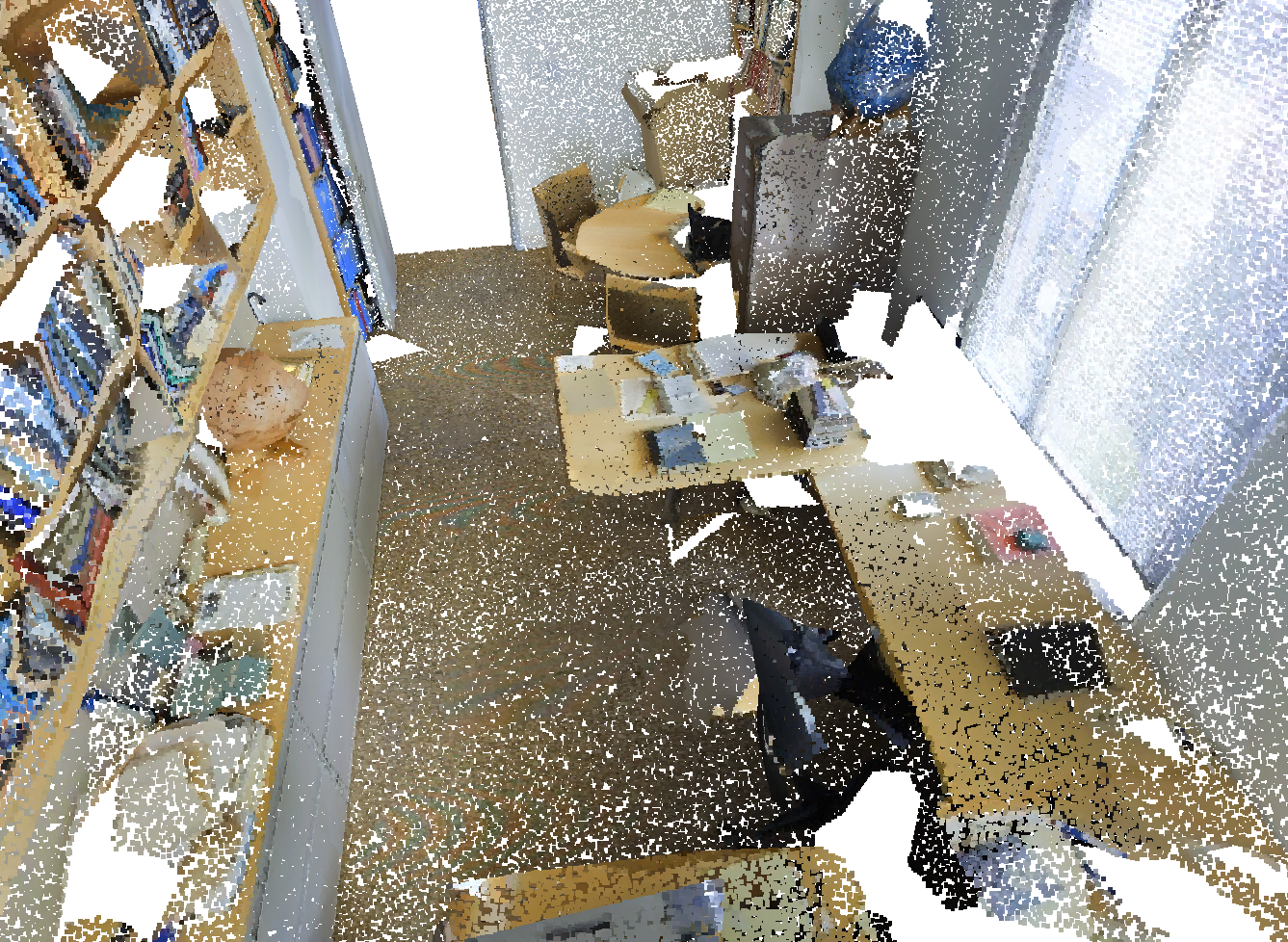}
        \caption{A raw sample}
    \end{subfigure}
    \hspace{10mm}
    \begin{subfigure}[b]{0.2\textwidth}
        \includegraphics[width=\textwidth]{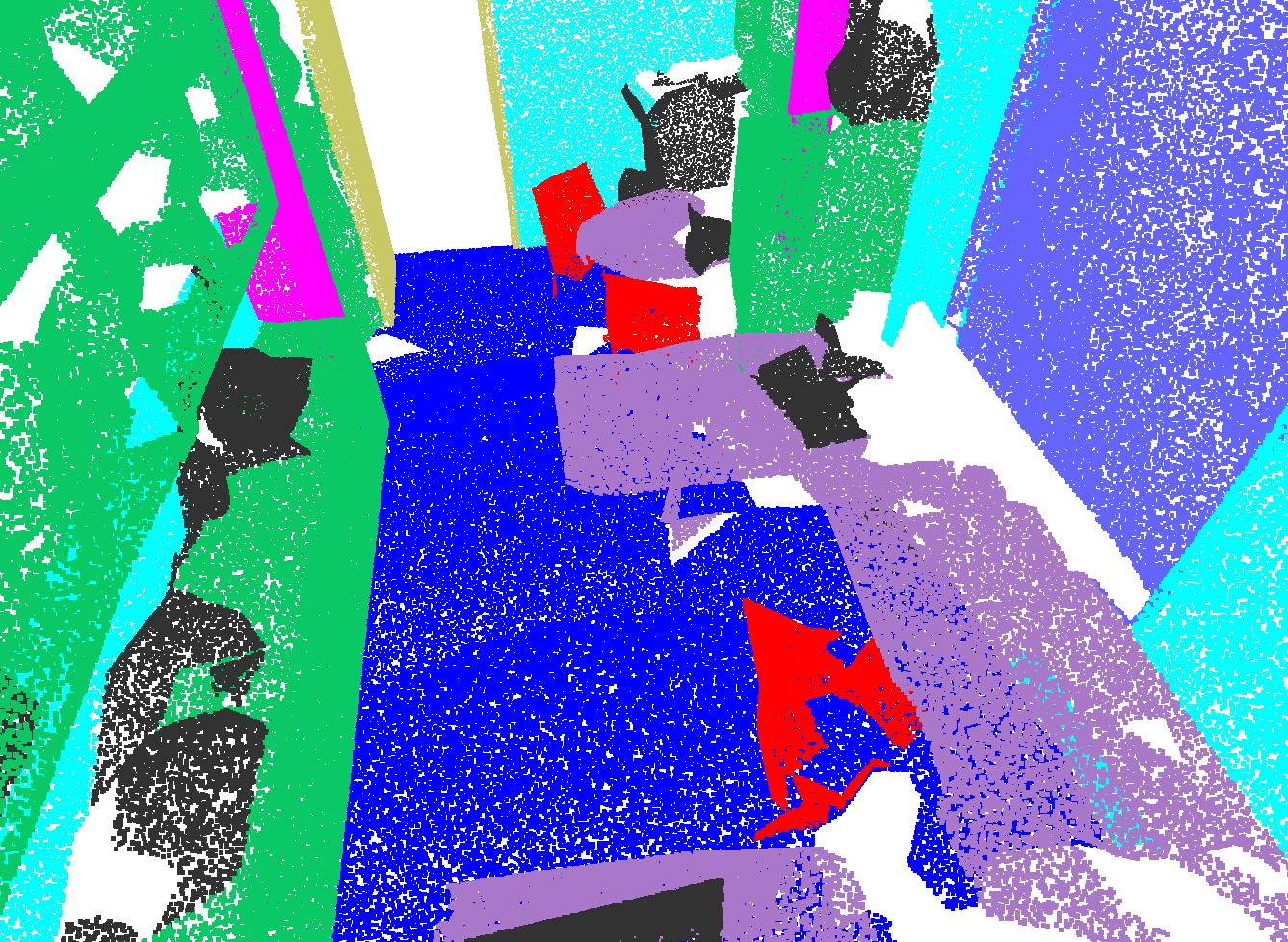}
        \caption{Semantic annotations}
    \end{subfigure}
    \caption{Illustration of semantic point cloud segmentation: For the point cloud sample from S3DIS \cite{armeni2016s3dis} on the left, the graph on the right shows the corresponding ground truth where different categories are highlighted by different colors.
    }
    \label{fig.semantic segmentation}
\end{figure}

\subsubsection{3D semantic segmentation}

3D semantic segmentation on point clouds is another critical task for 3D understanding as illustrated in Fig. \ref{fig.semantic segmentation}. Different from the object part segmentation that segments point cloud objects, 3D semantic segmentation aims to assign a category label to each point in scene-level point clouds with much higher complexity. The widely adopted evaluation metrics includes OA, mIoU over semantic categories and mAcc.

\subsubsection{3D instance segmentation}

\begin{figure}[h]
    \centering
    \begin{subfigure}[b]{0.23\textwidth}
        \includegraphics[width=\textwidth]{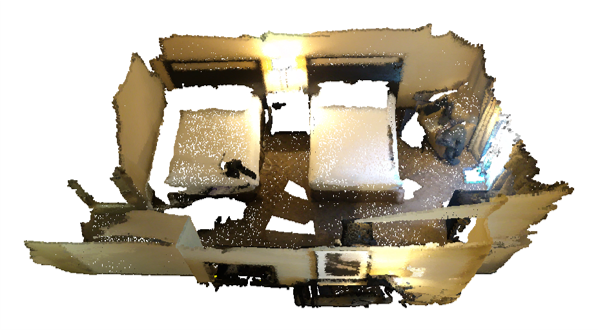}
        \caption{A raw sample}
    \end{subfigure}
    \begin{subfigure}[b]{0.23\textwidth}
        \includegraphics[width=\textwidth]{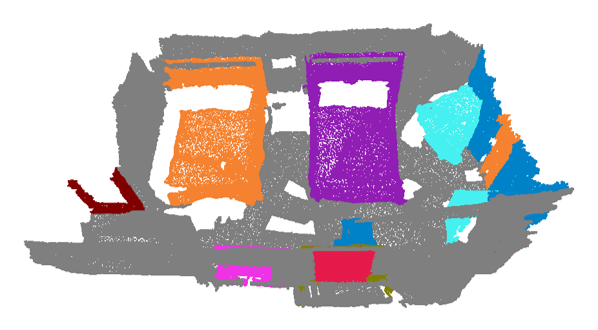}
        \caption{Instance annotations}
    \end{subfigure}
    \caption{Illustration of instance segmentation on point clouds: For the point cloud sample from ScanNet-V2~\cite{dai2017scannet} on the left, the graph on the right shows the corresponding ground truth with different instances highlighted by different colors.
    }
    \label{fig.instance segmentation}
\end{figure}

3D instance segmentation aims to detect and delineate each distinct object of interest in scene-level point clouds as illustrated in Fig. \ref{fig.instance segmentation}. On top of semantic segmentation that considers the semantic category only, instance segmentation assigns each object a unique identity. Mean Average Precision (mAP) has been widely adopted for the quantitative  evaluation of this task.

\subsection{Relevant surveys}
To the best of our knowledge, this paper is the \textit{first} survey that reviews unsupervised point cloud learning comprehensively. Several relevant but different surveys have been performed. For example, several papers reviewed recent advances for deep supervised learning on point clouds:
Ioannidou \textit{et al.}~\cite{ioannidou2017deep} reviewed deep learning approaches on 3D data;
Xie \textit{et al.}~\cite{xie2020linking} provided a literature review on point cloud segmentation task;
Guo \textit{et al.}~\cite{guo2020deep} provided a comprehensive and detailed survey on deep learning of point cloud for multiple tasks including classification, detection, tracking, and segmentation. In addition, several works reviewed unsupervised representation learning on other data modalities:
Jing \textit{et al.}~\cite{jing2020self} introduced advances on unsupervised representation learning in 2D computer vision; 
Liu \textit{et al.}~\cite{liu2021self} looked into latest progress about unsupervised representation learning methods in 2D computer vision, NLP, and graph learning;
Qi \textit{et al.}~\cite{qi2020small} introduced recent progress on small data learning including unsupervised- and semi-supervised methods.

\section{Point cloud datasets}\label{sec.PC_dataset}

\begin{table*}[t]
    \centering
    \caption{Summary of commonly used datasets for training and evaluations in prior URL studies with point clouds.}
    \begin{tabular}{|r|c|c|c|c|c|c|}
        \hline
        Dataset & Year & \#Samples & \#Classes & Type & Representation & Label\\
        \hline
        KITTI~\cite{geiger2013vision} & 2013 & 15K frames & 8 & Outdoor driving & RGB \& LiDAR & Bounding box\\
        ModelNet10~\cite{wu2015modelnet} & 2015 & 4,899 objects & 10 & Synthetic object & Mesh & Object category label\\
        ModelNet40~\cite{wu2015modelnet} & 2015 & 12,311 objects & 40 & Synthetic object & Mesh & Object category label\\
        ShapeNet~\cite{chang2015shapenet} & 2015 & 51,190 objects & 55 & Synthetic object & Mesh & Object/part category label\\
        SUN RGB-D~\cite{song2015sun} & 2015 & 5K frames & 37 & Indoor scene & RGB-D & Bounding box\\
        S3DIS~\cite{armeni2016s3dis} & 2016 & 272 scans & 13 & Indoor scene & RGB-D & Point category label\\
        ScanNet~\cite{dai2017scannet} & 2017 & 1,513 scans & 20 & Indoor scene & RGB-D \& mesh & Point category label \& Bounding box\\
        ScanObjectNN~\cite{uy2019scanobjectnn} & 2019 & 2,902 objects & 15 & Real-world object & Points & Object category label\\
        ONCE~\cite{mao2021one} & 2021 & 1M scenes & 5 & Outdoor driving & RGB \& LiDAR & Bounding box\\
        \hline
    \end{tabular}
    \label{tab.point cloud datasets}
\end{table*}

In this section, we summarize the commonly used datasets for training and evaluating unsupervised point cloud representation learning. As listed in Table~\ref{tab.point cloud datasets}, existing work learns unsupervised point cloud representations mainly from 1) synthetic object datasets including ModelNet~\cite{wu2015modelnet} and ShapeNet~\cite{chang2015shapenet}, or 2) real scene datasets including ScanNet~\cite{dai2017scannet} and KITTI~\cite{geiger2013vision}. 
In addition, various tasks-specific datasets have been collected which can be used for fine-tuning downstream models, such as ScanObjectNN~\cite{uy2019scanobjectnn}, ModelNet40~\cite{wu2015modelnet}, and ShapeNet~\cite{chang2015shapenet} for point cloud classification, ShapeNetPart~\cite{chang2015shapenet} for part segmentation, S3DIS~\cite{armeni2016s3dis}, ScanNet~\cite{dai2017scannet}, or Synthia4D~\cite{ros2016synthia} for semantic segmentation, indoor datasets SUNRGB-D~\cite{song2015sun} and ScanNet~\cite{dai2017scannet} as well as outdoor dataset ONCE~\cite{mao2021one} for object detection.

\noindent$\bullet$\textbf{ModelNet10/ModelNet40~\cite{wu2015modelnet}:} ModelNet is a synthetic object-level dataset for 3D classification. The original ModelNet provides CAD models represented by vertices and faces. Point clouds are generated by sampling from the models uniformly. 
    ModelNet40 contains 13,834 objects of 40 categories, among which 9,843 objects form the training set and the rest form the test set. ModelNet10 consists of 3,377 samples of 10 categories, which are split into 2,468 training samples and 909 testing samples. 
    
\noindent$\bullet$\textbf{ShapeNet~\cite{chang2015shapenet}:} ShapeNet contains synthetic 3D objects of 55 categories. It was curated by collecting CAD models from online open-sourced 3D repositories. Similar to ModelNet, synthetic objects in ShapeNet are complete, aligned, and with no occlusion or background. Its extension \textbf{ShapeNetPart} has 16,881 objects of 16 categories and is represented by point clouds. Each object consists of 2 to 6 parts, and in total there are 50 part  categories in the dataset.

\noindent$\bullet$\textbf{ScanObjectNN~\cite{uy2019scanobjectnn}:}    ScanObjectNN is a real  object-level dataset, where 2,902 3D point cloud objects of 15 categories are constructed from the scans captured in real indoor scenes. Different from synthetic object datasets, point cloud objects in ScanObjectNN are noisy (including background points, occlusions, and holes in objects) and not axis-aligned.

\noindent$\bullet$\textbf{S3DIS~\cite{armeni2016s3dis}:} Stanford Large-Scale 3D Indoor Spaces (S3DIS) dataset contains over 215 million points scanned from 6 large-scale indoor areas in 3 office buildings, where each area is 6,000 square meters. The scans are represented as point clouds with point-wise semantic labels of 13 object categories.

\noindent$\bullet$\textbf{ScanNet-V2~\cite{dai2017scannet}:} ScanNet-V2 is an RGB-D video dataset containing 2.5 million views in more than 1500 scans, which are captured in indoor scenes such as offices and living rooms and annotated with 3D camera poses, surface reconstructions, as well as semantic and instance labels for segmentation.

\noindent$\bullet$\textbf{SUN RGB-D~\cite{song2015sun}:} SUN RGB-D dataset is a collection of single view RGB-D images collected from indoor environments. There are in total 10,335 RGB-D images annotated with amodal, and 3D oriented object bounding boxes of 37 categories.

\noindent$\bullet$\textbf{KITTI~\cite{geiger2013vision}:} KITTI is a pioneer outdoor dataset providing dense point clouds from a LiDAR sensor together with other modalities including front-facing stereo images and GPS/IMU data. It provides 200k 3D boxes over 22 scenes for 3D object detection.

\noindent$\bullet$\textbf{ONCE~\cite{mao2021one}:} ONCE dataset has 1 million LiDAR scenes and 7 million corresponding camera images. There are 581 sequences in total, where 560 sequences are unlabelled and used for unsupervised learning, and 10 sequences are annotated and used for testing. It provides an unsupervised learning benchmark for object detection in outdoor environments.

The publicly available datasets for URL of point clouds are still limited in both data size and scene variety, especially compared with the image and text datasets that have been used for 2D computer vision and NLP research. For example, there are 800 million words in BooksCorpus and 2,500 million words in English Wikipedia that is able to provide comprehensive data sources for unsupervised representation learning in NLP~\cite{devlin2018bert};
ImageNet \cite{deng2009imagenet} has more than 10 million images for unsupervised visual representation learning. Large-scale and high-quality point cloud data are highly demanded for future research on this topic, and we provide a detailed discussion of this issue in Section~\ref{Sec.Future}.

\section{Common deep architectures}\label{Sec.DNN}

Over the last decade, deep learning has been playing a more important role in point-cloud processing and understanding. This can be observed by the abundance of deep architectures that have been developed in recent years. Different from traditional 3D vision that transforms point clouds to structures like Octrees \cite{hornung2013octomap} or Hashed Voxel Lists \cite{niessner2013real}, deep learning favors more amenable structures for differentiability and/or efficient neural processing which have achieved very impressive performance over various 3D tasks.

At the other end, DNN-based point cloud processing and understanding lags far behind as compared with its counterparts in NLP and 2D computer vision. This is especially true for the task of unsupervised representation learning, largely due to the lack of regular representations in point cloud data. Specifically, word embeddings and 2D images have regular and well-defined structures, but point clouds represented by unordered point sets have no such universal and structural data format.

In this section, we introduce deep architectures that have been explored for the URL of point clouds. Deep learning for point clouds achieved significant progress during the last decade and we see the abundance of 3D deep architectures and 3D models being proposed. However, we do not have universal and ubiquitous ``3D backbones” like VGG~\cite{Simonyan15} or ResNet~\cite{he2016deep} in 2D computer vision. We thus focus on those frequently used architectures in the URL of point clouds in this survey. For clarity of description, we group them into five categories broadly, namely, point-based architectures, graph-based architectures, sparse voxel-based architectures, spatial CNN-based architectures, and Transformer-based architectures. Note other deep architectures also exist for various 3D tasks as discussed in \cite{guo2020deep}, such as projection-based networks \cite{su2015multi,yu2018multi,yang2018pixor,yang2019learning,wei2020view,xiao2021fps}, recurrent neural networks \cite{huang2018recurrent,ye20183d,zou20173d}, 3D capsule networks \cite{zhao20193d}, etc. However, they were not often employed for the URL task and thus are not detailed in this survey.

\subsection{Point-based deep architectures}

Point-based networks were designed to process raw point clouds directly without point data transformations beforehand. Independent point features are usually first extracted by stacking networks with Multi-Layer Perceptrons (MLPs), which are then aggregated into global features with symmetric aggregation functions.

\begin{figure}[h]
    \centering
    \includegraphics[width=0.35\textwidth]{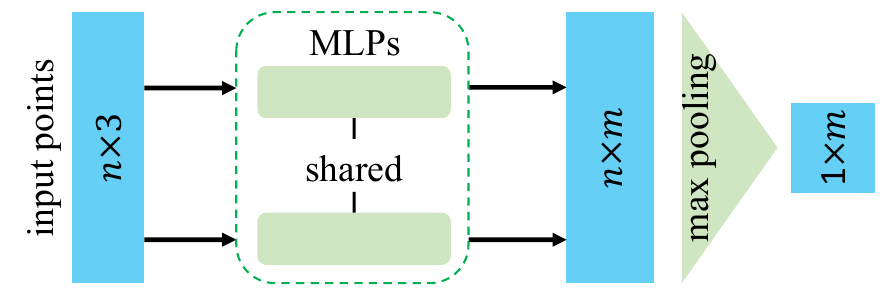}
    \caption{A simplified architecture of PointNet~\cite{qi2017pointnet} for point cloud object classification, where parameters $n$ and $m$ denote point number and feature dimension, respectively. }
    \label{fig:pointnet}
\end{figure}

PointNet~\cite{qi2017pointnet} is a pioneer point-based network as shown in Fig. \ref{fig:pointnet}. It stacks several MLP layers to learn point-wise features independently and forwards the learned features to a max-pooling layer to extract global features for permutation invariance. To improve PointNet, Qi \textit{et al.} proposed PointNet++ \cite{qi2017pointnet++} to learn local geometry details from the neighborhood of points, where the set abstraction level includes sampling layer, grouping layer, and PointNet layer for learning local and hierarchical features. PointNet++ achieves great success in multiple 3D tasks including object classification and semantic segmentation. By taking PointNet++ as the backbone, Qi \textit{et al.} designed VoteNet~\cite{qi2019votenet}, the first point-based 3D object detection network. VoteNet adopts the Hough voting strategy, which generates new points around object centers and groups them with the surrounding points to produce 3D box proposals.

\subsection{Graph-based deep architectures}

\begin{figure}[t]
    \centering
    \includegraphics[width=0.45\textwidth]{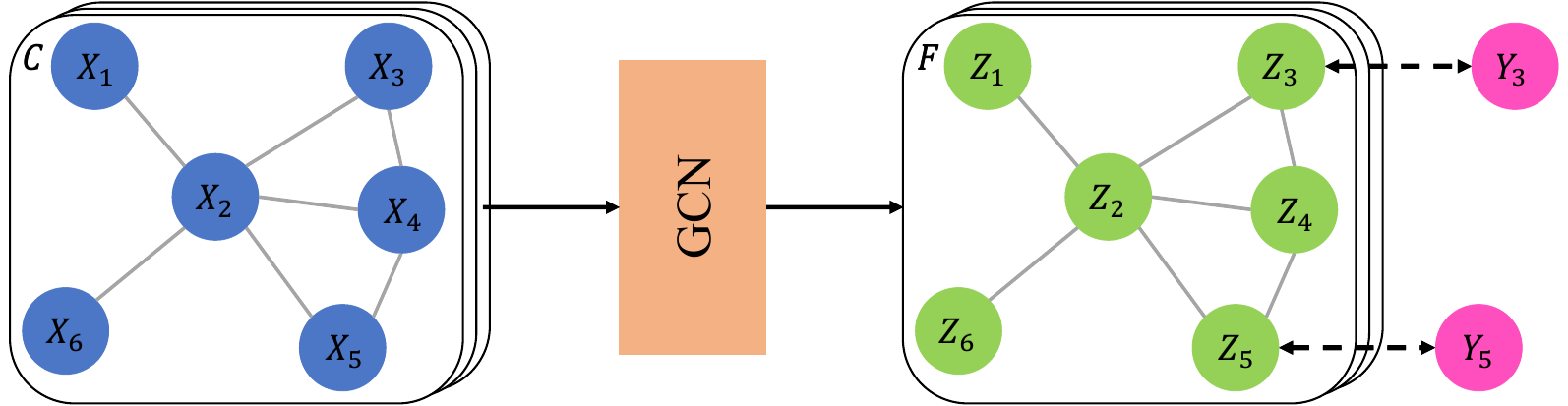}
    \caption{Schematic depiction of graph convolutional network (GCN): Each graph consists of multiple vertexes representing points $X_i$ or features $Z_i$ (highlighted by circular dots), as well as edges connecting the vertexes representing point relations (shown as black lines). $C$ denotes input channels, $F$ denotes output feature dimensions, and $Y_i$ denotes labels.}
    \label{fig:GCN}
\end{figure}

Graph-based networks treat point clouds as graphs in Euclidean space with vertexes being points and edges capturing neighboring point relations as illustrated in Fig. \ref{fig:GCN}. It works with graph convolution where filter weights are conditioned on edge labels and dynamically generated for individual input samples. This allows to reduce the degrees of freedom in the learned models by enforcing weight sharing and extracting localized features that can capture dependencies among neighboring points.

The Dynamic Graph Convolutional Neural Network (DGCNN)~\cite{wang2019dynamic} is a typical graph-based network that has been frequently used for URL for point clouds. It is stacked with a graph convolution module named EdgeConv that performs convolution on graph dynamically in the feature space. DGCNN integrates EdgeConv into the basic version of PointNet structures for learning  global shape properties and semantic characteristics for point cloud understanding.

\subsection{Sparse voxel-based deep architectures}

The voxel-based architecture voxelizes point clouds into 3D grids before applying 3D CNN on the volumetric representations. Due to the sparseness of point cloud data, It often involves huge computation redundancy or sacrifices the representation accuracy while processing a large number of points. To overcome this constrain, ~\cite{graham20183d,choy20194d,tang2022torchsparse,tang2020searching} adopt \textit{sparse tensor} as the basic unit where point clouds are represented with a data list and an index list. Unlike standard convolution operation that employs sliding windows (\textit{im2col} function in PyTorch and TensorFlow) to build the computational pipeline, \textit{sparse convolution}~\cite{graham20183d} collects all atomic operations including convolution kernel elements and saves them in a \textit{Rulebook} as computation instructions.

\begin{figure}[b]
    \centering
    \includegraphics[width=0.5\textwidth]{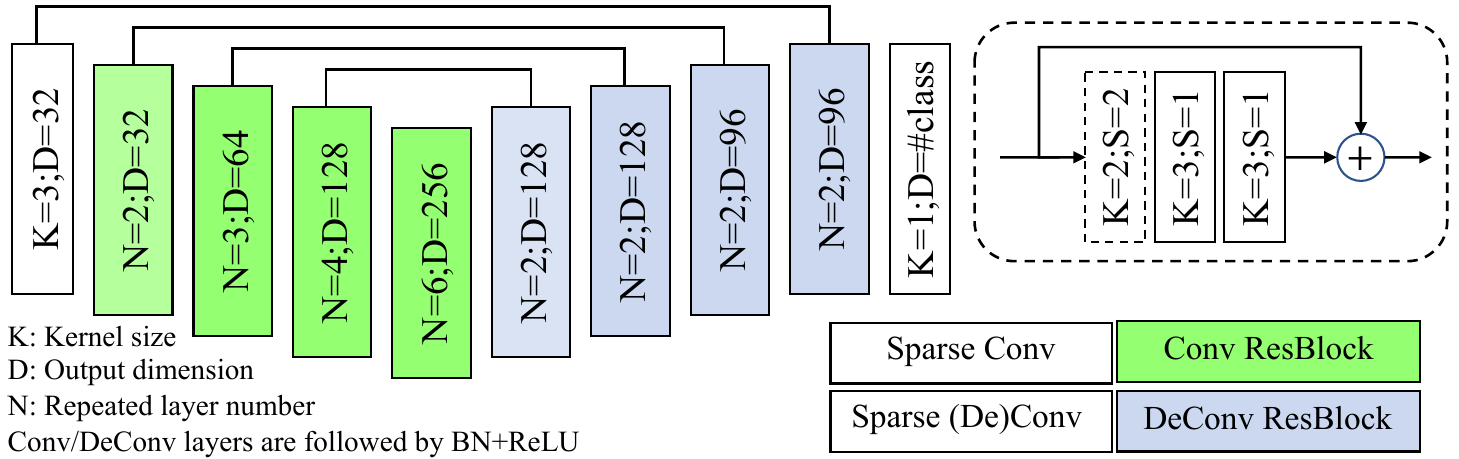}
    \caption{An illustration of SR-UNet~\cite{xie2020pointcontrast} that adopts a unified U-Net~\cite{ronneberger2015u} architecture for sparse convolution. The graph is reproduced based on \cite{xie2020pointcontrast}.}
    \label{fig:sr-unet}
\end{figure}

Recently, Choy et al. proposed Minkowski Engine \cite{choy20194d} that introduces generalized sparse convolution and an auto-differentiation library for sparse tensors. On top of that, Xie et al. \cite{xie2020pointcontrast} adopted a unified U-Net~\cite{ronneberger2015u} architecture and built a backbone network (SR-UNet as shown in Fig. \ref{fig:sr-unet}) for unsupervised pre-training. The learned encoder can be transferred to different downstream tasks such as classification, object detection, and semantic segmentation.

\subsection{Spatial CNN-based deep architectures}
\begin{figure}[t]
    \centering
    \includegraphics[width=0.45\textwidth]{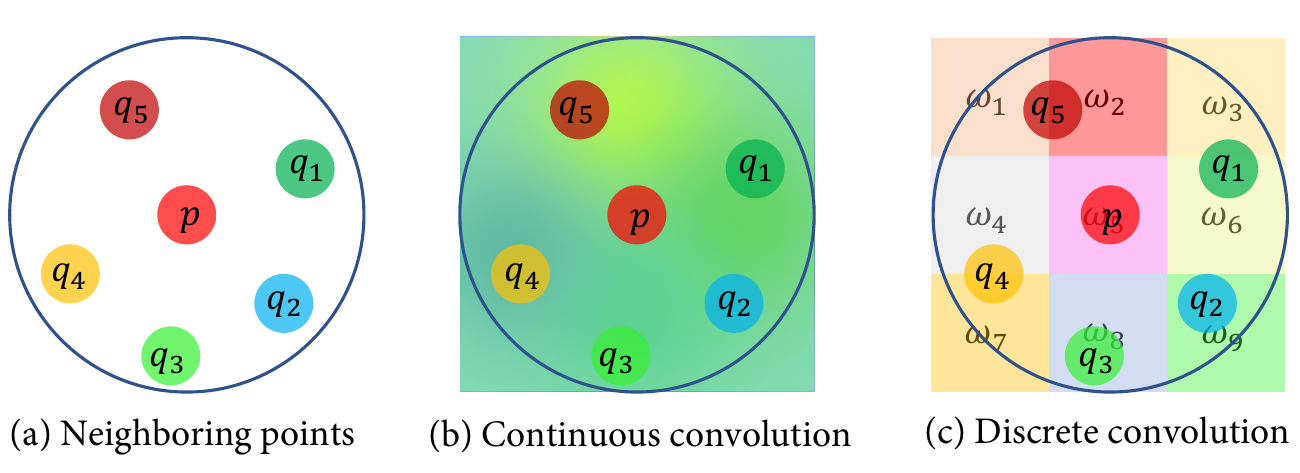}    
    \caption{An illustration of 3D spatial convolution including continuous and discrete convolutions. Parameters $p$ and $q_i$ denote the center point and its neighboring points, respectively. The graph is reproduced based on \cite{guo2020deep}.
    }
    \label{fig:3dconv}
\end{figure}

Spatial CNN-based networks have been developed to extend the capabilities of regular-grid CNNs to analyze irregularly spaced point clouds.
They can be divided into continuous and discrete convolutional networks according to the convolutional kernels \cite{guo2020deep}. As Fig.~\ref{fig:3dconv} shows, continuous convolutional networks define the convolutional kernels in a continuous space, where the weights of neighboring points are determined by their spatial distribution relative to the center point. Differently, discrete convolutional networks operate on regular grids and define the convolutional kernels in a discrete space where neighboring points have fixed offsets relative to the center point.
One typical example of continuous convolution models is RS-CNN~\cite{liu2019rscnn} which has been widely adopted for URL of point clouds. Specifically, RS-CNN extracts geometric topology relations among local centers with their surrounding points, and it learns dynamic weights for convolutions.

\subsection{Transformer-based deep architectures}

\begin{figure}[b]
    \centering
    \includegraphics[width=0.45\textwidth]{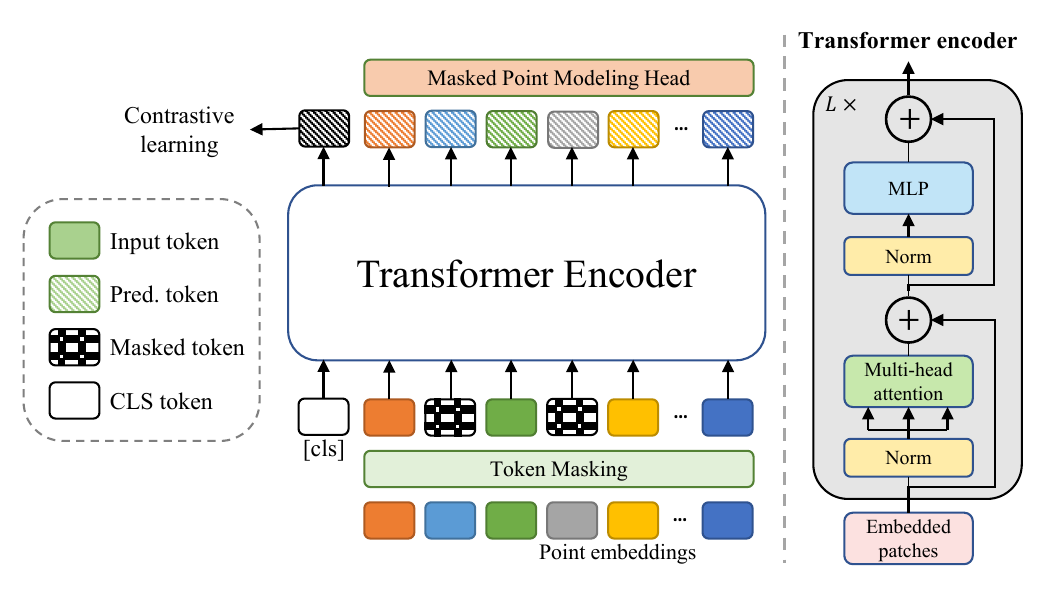}    
    \caption{The architecture of point cloud Transformer that was used for unsupervised pre-training in Point-BERT~\cite{yu2022point}. More network details can be found in \cite{yu2022point}. The figure is reproduced based on \cite{yu2022point,dosovitskiy2020vit}.}
    \label{fig:transformer}
\end{figure}

Over the last few years, Transformers have made astounding progress in the research areas of NLP~\cite{vaswani2017attention,devlin2018bert} and 2D image processing~\cite{dosovitskiy2020vit,liu2021swin} due to their structural superiority and versatility. They have also been introduced into the area of point cloud processing~\cite{Zhao_2021_ICCV,yu2022point} recently. Fig.~\ref{fig:transformer} shows a standard Transformer architecture for URL of point clouds~\cite{yu2022point}, which contains a stack of Transformer blocks~\cite{vaswani2017attention} and each block consists of a multi-head self-attention layer and a feed-forward network. The unsupervised pre-trained Transformer encoder can be used for fine-tuning downstream tasks such as object classification and semantic segmentation, etc.

\section{Unsupervised point cloud representation learning}\label{Sec.URL}

In this section, we review existing URL methods for point clouds. As shown in Fig. \ref{fig.landscape_PC-URL}, we broadly group existing methods into four categories according to their pretext tasks, including generative-based methods, context-based methods, multiple modal-based methods, and local descriptor-based methods.
With this taxonomy, we sort out existing methods and systematically introduce them in the ensuing subsections of this section.

\subsection{Generation-based methods}\label{Subsec.Generation}

\begin{table*}[h]
    \setlength\tabcolsep{6pt}
    \centering
    \caption{Summary of generation-based methods for unsupervised representation learning of point clouds.}
    \begin{tabular}{|l|c|c|l|}
    \hline
        Method & Published in & Category & Contribution \\
    \hline
        VConv-DAE~\cite{sharma2016vconv-dae} & ECCV 2016 & Completion & Learning by predicting missing parts in 3D grids\\
        TL-Net~\cite{girdhar2016tl_net} & ECCV 2016 & Reconstruction & Learning by 3D generation and 2D prediction\\
        3D-GAN~\cite{wu20163d-gan} & NeurIPS 2016 & GAN & Pioneer GAN for 3D voxels\\
        3D-DescriptorNet \cite{xie2018learning} & CVPR 2018 & Completion & learning with energy-based models for point cloud completion\\
        FoldingNet~\cite{yang2018foldingnet} & CVPR 2018 & Reconstruction & learning by folding 3D object surfaces\\
        SO-Net~\cite{li2018so-net} & CVPR 2018 & Reconstruction & Performing hierarchical feature extraction on individual points and SOM nodes\\
        Latent-GAN~\cite{achlioptas2018learning} &  ICML 2018 & GAN  & Pioneer GAN for raw point clouds and latent embeddings\\
        MRT~\cite{gadelha2018multiresolution} & ECCV 2018 & Reconstruction & A new point cloud autoencoder with multi-grid architecture\\
        VIP-GAN~\cite{han2019view} & AAAI 2019 & GAN & Learning by solving multi-views inter-prediction tasks for objects \\ 
        G-GAN~\cite{valsesia2018learning} & ICLR 2019 & GAN & Pioneer GAN with graph convolution for point clouds\\
        3DCapsuleNet~\cite{zhao20193d} & CVPR 2019 & Reconstruction & Learning with 3D point-capsule network \\
        L2G-AE~\cite{liu2019l2g} & ACM MM 2019 & Reconstruction & Learning by global and local reconstruction of point clouds\\
        MAP-VAE~\cite{han2019multi} & ICCV 2019 & Reconstruction & Learning by 3D reconstruction and half-to-half prediction\\
        PointFlow~\cite{yang2019pointflow} & ICCV 2019 & Reconstruction & Learning by modeling point clouds as a distribution of distributions\\
        PDL~\cite{shi2020unsupervised} & CVPR 2020 & reconstruction & A probabilistic framework for point distribution learning\\
        GraphTER~\cite{gao2020graphter} & CVPR 2020 &  Reconstruction & Proposed a graph-based autoencoder for point clouds\\
        SA-Net~\cite{wen2020point} & CVPR 2020 & Completion & Learning by completing point cloud objects with a skip-attention mechanism\\
        PointGrow~\cite{sun2020pointgrow} & WACV 2020& Reconstruction & An autoregressive model that can recurrently generate point cloud samples.\\
        PSG-Net~\cite{yang2021progressive} & ICCV 2021 & Reconstruction & Learning by reconstruct point cloud objects with seed generation \\
        OcCo~\cite{Wang_2021_ICCV} & ICCV 2021 & Completion & Learning by completing occluded point cloud objects\\
        Point-Bert~\cite{yu2022point} & CVPR 2022 & Reconstruction & Learning for Transformers by recovering masked tokens of 3D objects\\
        Point-MAE~\cite{pang2022masked} & ECCV 2022 & Reconstruction & Autoencoder transformer recovers masked parts from input data\\
        Point-M2AE~\cite{zhang2022point} & NeurIPS 2022 & Reconstruction & Masked autoencoder with hierarchical point cloud encoding and reconstruction. \\
    \hline
    \end{tabular}
    \label{tab. Sum of generative methods}
\end{table*}

Generation-based URL methods for point clouds involve the process of generating point cloud objects in training. According to the employed pre-text tasks, they can be further grouped into four subcategories including point cloud self-reconstruction (for generating point cloud objects that are the same as the input), point cloud GAN (for generating fake point cloud objects), point cloud up-sampling (for generating objects with denser point clouds but similar shapes) and point cloud completion (for predicting missing parts from incomplete point cloud objects). The ground truth of these URL methods are point clouds themselves. Hence, these methods require no human annotations and can learn in an unsupervised manner. Table~\ref{tab. Sum of generative methods} shows a list of generation-based methods.

\subsubsection{Learning through point cloud self-reconstruction}\label{sec.self-reconstruct}

\begin{figure}[b]
    \centering
    \includegraphics[width=0.5\textwidth]{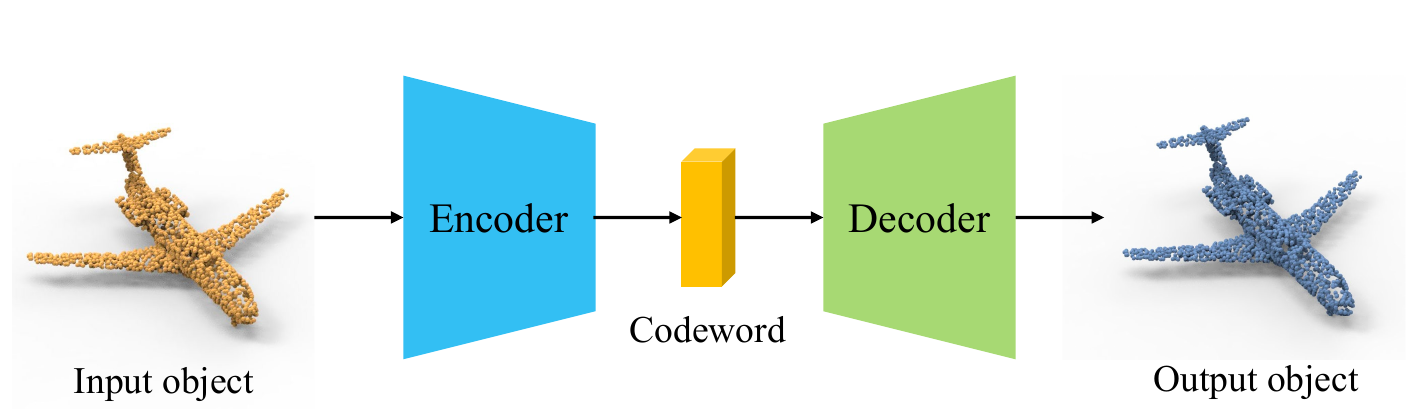}
    \caption{An illustration of AutoEncoder in unsupervised point cloud representation learning: The \textit{Encoder} learns to represent a point cloud object by a \textit{Codeword} vector while the \textit{Decoder} reconstructs the \textit{Output Object} from the \textit{Codeword}.}
    \label{fig. autoencoder}
\end{figure}

Networks for self-reconstruction usually encode point cloud samples into representation vectors and decode them back to the original input data, where shape information and semantic structures are extracted during this process. It belongs to one typical URL approach since it does not involve any human annotations. One representative network is \textit{autoencoder} \cite{kramer1991nonlinear} which has an encoder network and a decoder network as illustrated in Fig. \ref{fig. autoencoder}. The encoder compresses and encodes a point cloud object into a low-dimensional embedding vector (i.e., \textit{codeword})~\cite{yang2018foldingnet}, which is then decoded back to the 3D space by the decoder.

The model is optimized by forcing the final output to be the same as the input. During this process, the encoding is validated and learns by attempting to regenerate the input from the encoding whereas the autoencoder learns low-dimension representations by training the network to ignore insignificant data (“noise”) \cite{wiki:Autoencoder}. Permutation invariant losses \cite{fan2017point} are widely adopted as the training objective to describe how the input and output point cloud objects are similar to each other. They can be measured by  Chamfer Distance $L_{\mathrm{CD}}$ or Earth Mover’s Distance $L_{\mathrm{EMD}}$ as follows:
\begin{equation}
    L_{\mathrm{CD}} = \sum_{p\in P} \min_{p'\in P'} {||p-p'||}^2 + \sum_{p'\in P'} \min_{p\in P} {||p'-p||}^2
    \label{eq. chamfer}
\end{equation}
\begin{equation}
    L_{\mathrm{EMD}} = \min_{\phi:P\rightarrow P'} \sum_{x\in P} {||p-\phi(p)_2||}_2
    \label{eq. EMD}
\end{equation}
Where $P$ and $P'$ denote input and output point clouds of the same size, $\phi:P\rightarrow P'$ is bijection, and $p$ \& $p'$ are points.

Self-reconstruction has been one of the most widely adopted pre-text tasks for URL from point clouds over the last decade.  By assuming that point cloud representations should be generative in 3D space and predictable from 2D space, Girdhar \textit{et al.} proposed TL-Net~\cite{girdhar2016tl_net} that employs a 3D autoencoder to reconstruct 3D volumetric grids and a 2D convolutional network to learn 2D features from the projected images. Yang \textit{et al.} designed FoldingNet~\cite{yang2018foldingnet} that introduces a folding-based decoder that deforms a canonical 2D grid onto the underlying 3D object surface of a point cloud object. Li \textit{et al.} proposed SO-Net~\cite{li2018so-net} that introduces self-organizing map to learn hierarchical features of point clouds via self-reconstruction. Zhao \textit{et al.}~\cite{zhao20193d} extended the capsule network~\cite{sabour2017dynamic} into 3D point cloud processing and the designed 3D capsule network can learn generic representations from unstructured 3D data. Gao \textit{et al.}~\cite{gao2020graphter} proposed a graph-based autoencoder that can learn intrinsic patterns of point-cloud structures under both global and local transformations. Chen \textit{et al.}~\cite{chen2020deep} designed a deep autoencoder that exploits graph topology inference and filtering for extracting compact representations from 3D point clouds.

Several studies explore global and local geometries to learn robust representations from point cloud objects~\cite{liu2019l2g,han2019multi}. For example, \cite{liu2019l2g} introduces hierarchical self-attention in the encoder for information aggregation, and a recurrent neural network (RNN) as the decoder for point cloud reconstruction locally and globally. \cite{han2019multi} presents MAP-VAE that introduces a half-to-half prediction task that first splits a point cloud object into a front half and a back half and then trains an RNN to predict the back half sequence from the corresponding front half sequence. Several studies instead formulate point cloud reconstruction as a point distribution learning task \cite{yang2019pointflow, shi2020unsupervised, sun2020pointgrow}. For example, \cite{yang2019pointflow} presents PointFlow which generates 3D point clouds by modelling the distribution of shapes and that of points given shapes. \cite{shi2020unsupervised} presents a probabilistic framework that extracts unsupervised shape descriptors via point distribution learning, which associates each point with a Gaussian and models point clouds as the distribution of points. \cite{sun2020pointgrow} presents an autoregressive model Pointgrow that generates diverse and realistic point cloud samples either from scratch or conditioned on semantic contexts.

Further, several studies learn point cloud representations from different object resolutions \cite{gadelha2018multiresolution,yang2021progressive,chen2021unsupervised}. For example, Gadelha \textit{et al.}~\cite{gadelha2018multiresolution} designed an autoencoder with a multi-resolution tree structure that learns point cloud representations via coarse-to-fine analysis. Yang \textit{et al.}~\cite{yang2021progressive} proposed an autoencoder with a seed generation module that allows extraction of input-dependent point-wise features in multiple stages with gradually increasing resolution. Chen \textit{et al.}~\cite{chen2021unsupervised} proposed to learn sampling-invariant features by reconstructing point cloud objects of different resolutions and minimizing Chamfer distances between them.

\begin{figure}[t]
    \centering
    \includegraphics[width=0.5\textwidth]{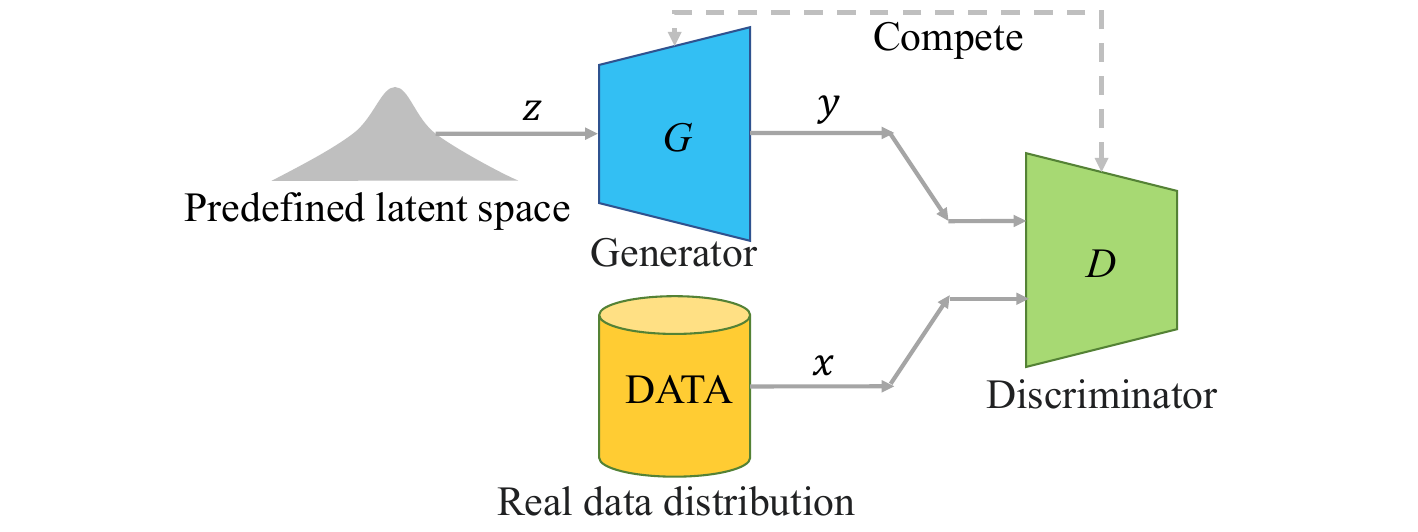}
    \caption{An illustration of GAN which typically consists of a generator $G$ and a discriminator $D$ that fight with each other during the training process (in the form of a zero-sum game, where one agent's gain is another agent's loss).}
    \label{fig. gan}
\end{figure}

\subsubsection{Learning through point cloud GAN}\label{sec.gan}

Generative and Adversarial Network (GAN) \cite{goodfellow2014generative} is a typical deep generative network. As demonstrated in Fig. \ref{fig. gan}, it consists of a generator and a discriminator. The generator aims to synthesize as realistic data samples as possible while the discriminator tries to differentiate real samples and synthesized samples. GAN thus learns to generate new data with the same statistics as the training set and the modeling can be formulated as a min-max problem:
\begin{equation}
    \min_G \max_D L_{GAN} = \log D(x) + \log (1 - D(G(z))),
\end{equation}
where $G$ is the generator and $D$ represents the discriminator. $x$ and $z$ represent a real sample and a randomly sampled noise vector from a distribution $p(z)$, respectively.

When training GANs for URL of point clouds, the generator learns from either a sampled vector or a latent embedding to generate point cloud instances, while the discriminator tries to distinguish whether input point clouds are from real data distribution or generated data distribution. The two sub-networks fight with each other during the training process and the discriminator learns to extract useful feature representations for point cloud object recognition. The learning process involves no human annotations thus the networks can be trained in an unsupervised learning manner.
After that, the learned discriminator is extended into various downstream tasks such as object classification or part segmentation by fine-tuning the model. 

Several networks employ GAN for URL for point clouds successfully \cite{wu20163d-gan,achlioptas2018learning,valsesia2018learning,li2018point}. For example, Wu \textit{et al.}~\cite{wu20163d-gan} proposed the first GAN model applying for 3D voxels. However, the voxelization process either sacrifices the representation accuracy or incurs huge redundancies. Achlioptas \textit{et al.} proposed Latent-GAN~\cite{achlioptas2018learning} as the first GAN model for raw point clouds. Li \textit{et al.}~\cite{li2018point} further proposed a point cloud GAN model with a hierarchical sampling and inference network that learns a stochastic procedure to generate new point cloud objects. Valsesia \textit{et al.}~\cite{valsesia2018learning} designed the first graph-based GAN model to extract localized features from point clouds. These methods evaluated the generalization of the learned representations by fine-tuning them to the high-level downstream 3D tasks.

\subsubsection{Learning through point cloud up-sampling}

\begin{figure}[t]
    \centering
    \includegraphics[width=0.4\textwidth]{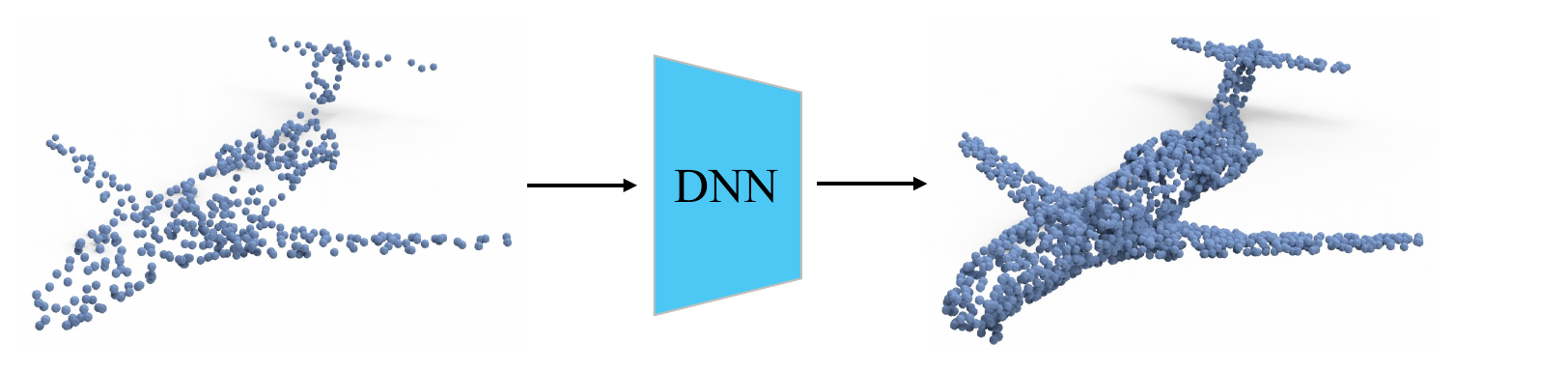}
    \caption{An illustration of point cloud up-sampling: The network \textit{DNN} learns point cloud representations by solving a pre-text task that reproduces an object with the same geometry but denser point distribution.}
    \label{fig. point cloud up-sampling}
\end{figure}

As shown in Fig. \ref{fig. point cloud up-sampling}, given a set of points, point cloud up-sampling aims to generate a denser set of points with similar geometries. This task requires deep point cloud networks to learn underlying geometries of 3D shapes without any supervision, and the learned representations can be used for fine-tuning in 3D downstream tasks.

Li \textit{et al.}~\cite{li2019pu} introduced GAN into the point cloud up-sampling task and presented PU-GAN to learn a variety of point distributions from the latent space by up-sampling points over patches on object surfaces. The generator aims to produce up-sampled point clouds while the discriminator tries to distinguish whether its input point cloud is produced by the generator or the real one. Similar to GANs introduced in Section~\ref{sec.gan}, the learned discriminator can be transferred in downstream tasks. Remelli \textit{et al.}~\cite{remelli2019neuralsampler} designed an autoencoder that can up-sample sparse point clouds into dense representations. The learned weight of the encoder can also be used as initialization weights for downstream tasks as described in Section~\ref{sec.self-reconstruct}. Though point cloud up-sampling is attracting increasing attention in recent years~\cite{yu2018pu,yifan2019patch,li2019pu,qian2020pugeo,qian2021pu,li2021point}, it is largely evaluated by the quality of generated point clouds while its performance in transfer learning has not been well studied.

\begin{figure}[b]
    \centering
    \includegraphics[width=0.45\textwidth]{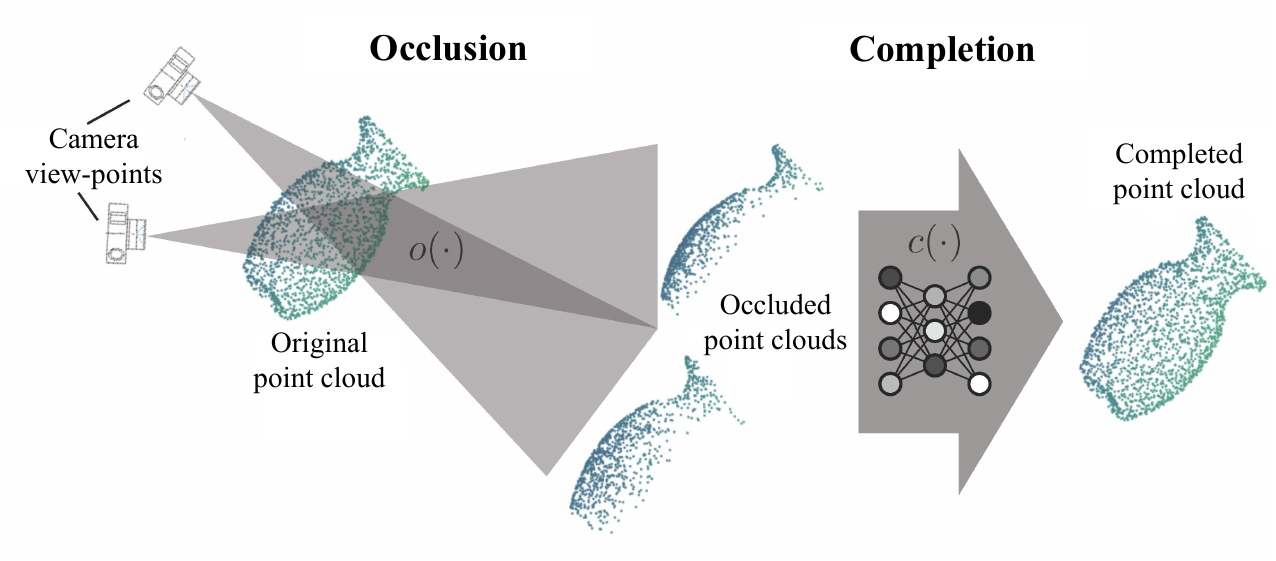}
    \caption{The pipeline of OcCo~\cite{Wang_2021_ICCV}. Taking occluded point cloud objects as input, an encoder-decoder model is trained to complete the occluded point clouds, where the encoder learns point cloud representations and the decoder learns to generate complete objects. The learned encoder weights can be used for network initialization for downstream tasks. The figure is from \cite{Wang_2021_ICCV} with author's permission.}
    \label{fig. point cloud completion}
\end{figure}

\subsubsection{Learning through point cloud completion}

\begin{table*}[h]
    \setlength\tabcolsep{4pt}
    \centering
    \caption{Summary of context-based methods for unsupervised representation learning of point clouds. }
    \begin{tabular}{|l|c|c|l|}
    \hline
        Method & Published in & Category & Contribution \\
    \hline
        MultiTask~\cite{hassani2019unsupervised} & ICCV 2019 & Hybrid & Learning by clustering, reconstruction, and self-supervised classification\\
        Jigsaw3D~\cite{sauder2019self} & NeurIPS 2019 & Spatial-context & Learning by solving 3D jigsaws\\
        Constrast\&Cluster~\cite{zhang2019unsupervised} & 3DV 2019 & Hybrid & Learning by contrasting and clustering with GNN\\
        GLR~\cite{rao2020global} & CVPR 2020 & Hybrid  & Learning by global-local reasoning for 3D objects\\
        Info3D~\cite{sanghi2020info3d} & ECCV 2020 & Context-similarity & Learning by contrasting global and local parts of objects\\
        PointContrast~\cite{xie2020pointcontrast} & ECCV 2020 & Context-similarity & Learning by contrasting different views of scene point clouds\\
        ACD~\cite{gadelha2020label} & ECCV 2020 & Context-similarity & Learning by contrasting  convex components decomposed from 3D objects\\
        Rotation3D~\cite{poursaeed2020self} & 3DV 2020 & Spatial-context & Learning by predicting rotation angles\\ 
        HNS~\cite{du2021self} & ACM MM 2021 & Context-similarity & Learning by contrasting local patches of 3D objects with hard negative sampling \\
        CSC~\cite{hou2021exploring} & CVPR 2021 & Context-similarity & Techniques to improve contrasting scene point cloud views\\
        STRL~\cite{huang2021spatio} & ICCV 2021 & Temporal-context & Learning  spatio-temporal data invariance from point cloud sequences\\
        RandomRooms~\cite{rao2021randomrooms} & ICCV 2021 & Context-similarity & Constructing pseudo scenes with synthetic objects for contrastive learning\\
        DepthContrast~\cite{Zhang_2021_ICCV} & ICCV 2021 & Context-similarity & Joint contrastive learning with points and voxels\\
        SelfCorrection~\cite{chen2021shape} & ICCV 2021 & Hybrid & Learning by distinguishing and restoring destroyed objects\\
        PC-FractalDB~\cite{Yamada_2022_CVPR} & CVPR 2022 & Context-similarity & Leveraging fractal geometry to generate high-quality pre-training data \\
        4dcontrast~\cite{chen20224dcontrast} & ECCV 2022 & Temporal-context & Learning by contrasting dynamic correspondences from 3D scene sequences \\
        DPCo~\cite{li2022closer} & ECCV 2022 & Context-similarity & A unified contrastive-learning framework for point cloud pre-training \\
        ProposalContrast~\cite{yin2022proposalcontrast} & ECCV 2022 & Context-similarity & Pre-training 3D detectors by contrasting region proposals \\
        MaskPoint~\cite{liu2022masked} & ECCV 2022 & Context-similarity & Learning by discriminating masked object points and sampled noise points\\
        FAC~\cite{liu2023fac} & CVPR 2023 & Context-similarity & Learning by contrasting between grouped foreground and background\\
    \hline
    \end{tabular}
    \label{tab. Sum of context methods}
\end{table*}

Point cloud completion is a task to predict arbitrary missing parts based on the rest of the 3D point clouds. To achieve this target, deep networks need to learn inner geometric structures and semantic knowledge of the 3D objects so as to correctly predict missing parts. On top of that, the learned representations can be transferred to downstream tasks. The whole process involves no human annotations and thus belongs to unsupervised representation learning.

Point cloud completion has been an active research area over the past decade~\cite{groueix2018papier,yuan2018pcn,PF-Net,liu2020morphing,wen2020point, zhang2020detail,xie2021style} with evaluation in different URL benchmarks  \cite{sharma2016vconv-dae,xie2018learning,wen2020point,Wang_2021_ICCV}. A pioneer work VConv-DAE \cite{sharma2016vconv-dae} voxelizes point cloud objects into volumetric grids and learns object shape distributions with an autoencoder by predicting the missing voxels from the rest parts. Xie \textit{et al.}~\cite{xie2018learning} designed 3D-DescriptorNet for probabilistic modeling of volumetric shape patterns. Achlioptas \textit{et al.}~\cite{achlioptas2018learning} introduced the first DNN for raw point cloud completion which is a point-based network with an encoder-decoder structure. Yuan \textit{et al.}~\cite{yuan2018pcn} proposed a Point Completion Network, an autoencoder structured network for learning useful representations by repairing incomplete point cloud objects. Wen \textit{et al.}~\cite{wen2020point} proposed SA-Net, which introduces a skip-attention mechanism in the encoder that selectively transfers geometric information from the local regions to the decoder for generating complete point cloud objects. Wang \textit{et al.}~\cite{Wang_2021_ICCV} proposed to learn an encoder-decoder model that recovers the occluded points by different camera views as shown in Fig. \ref{fig. point cloud completion}. The encoder parameters are used as initialization for downstream tasks including classification, part segmentation, and semantic segmentation.

Recently, recovering missing parts from \textit{masked} input as the pre-text task of URL has been proved remarkably successful in NLP~\cite{radford2019language,kenton2019bert} and 2D computer vision \cite{he2022masked}.  Such idea has also been investigated in 3D point cloud learning~\cite{yu2022point,pang2022masked,liu2022masked,fu2022pos}. For example, Yu \textit{et al.}~\cite{yu2022point} proposed a Point-BERT paradigm that pre-trains point cloud Transformers through a masked point modeling task. They use a discrete variational autoencoder to generate tokens for object patches and randomly masked out the tokens to train the Transformer to recover the original complete point tokens. The representations learned by Point-BERT can be well transferred to new tasks and domains such as object classification and object part segmentation.

\subsection{Context-based methods}\label{Sec.Context-based}

Context-based methods are another important category of URL of point clouds that has attracted increasing attention in recent years. Different from generation-based methods that learn representations in a generative way, these methods employ discriminative pre-text tasks to learn different contexts of point clouds including context similarity, spatial context structures, and temporal context structures. The designed pre-text tasks require no human annotations and Table~\ref{tab. Sum of context methods} lists the recent methods.

\subsubsection{Learning with context similarity}
\begin{figure}[b]
    \centering
    \includegraphics[width=0.5\textwidth]{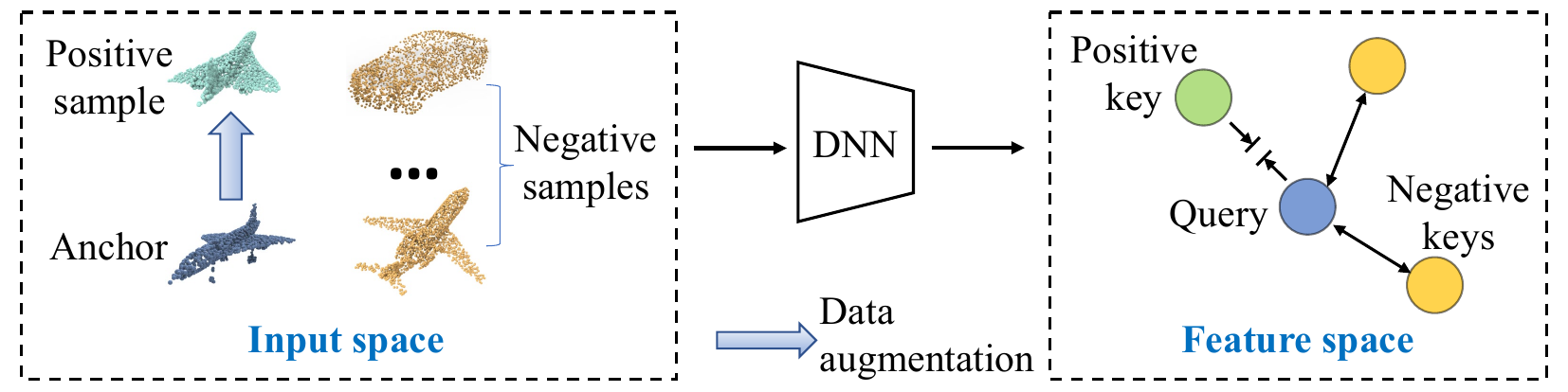}
    \caption{An illustration of instance contrastive learning that learns locally smooth representations by self-discrimination, which pulls \textit{Query} (from the \textit{Anchor} sample) close to \textit{Positive Key}  (from \textit{Positive Samples}) and pushes it away from \textit{Negative Keys}  (from \textit{Negative Samples}).}
    \label{fig.contrastive learning}
\end{figure}

This type of method learns unsupervised representations of point clouds by exploring underlying context similarities between samples. A typical approach is \textit{contrastive learning}, which has demonstrated superior performances in both 2D vision~\cite{he2020momentum,grill2020bootstrap,chen2020simple} and 3D vision~\cite{xie2020pointcontrast,hou2021exploring,Zhang_2021_ICCV} in recent years. Fig. \ref{fig.contrastive learning} provides an illustration of instance-wise contrastive learning. Given one input point cloud object instance as the anchor, its augmented views are defined as the positive samples while other different instances are negative samples. The network learns representations of point clouds by optimizing a self-discrimination task, \textit{i.e.} query (feature of the anchor) should be close to the positive keys (features of positive samples) and faraway from its negative keys (features of negative samples). This learning strategy groups representations of similar samples together in an unsupervised manner and helps networks to learn semantic structures from unlabelled data distribution. The InfoNCE loss~\cite{oord2018representation} defined below and its variants are often employed as the objective function in training:
\begin{equation}
    \mathcal{L}_{\mathrm{InfoNCE}}=-\log \frac{\exp{(q\cdot k_+/\tau)}}{\sum_{i=0}^{K}{\exp(q\cdot k_i/\tau)}},
\end{equation}
where $q$ is encoded query, $\{k_0, k_1, k_2, ...\}$ are keys with $k_+$ being the positive key, $\tau$ is a temperature hyper-parameter that controls how the distribution concentrates. 

Similar to generation-based methods, different contrastive learning methods \cite{sanghi2020info3d,wang2020unsupervised,gadelha2020label,du2021self,jiang2021unsupervised} have been proposed to learn representations on \textit{synthetic single objects}. For example, Sanghi \textit{et al.}~\cite{sanghi2020info3d} proposed to learn useful feature representations by maximizing mutual information between synthetic objects and their local parts.
Wang \textit{et al.}~\cite{wang2020unsupervised} proposed a hybrid contrastive learning strategy that uses objects of different resolutions for instance-level contrast for capturing hierarchical global representations and simultaneously contrasted points and instances for learning local features.
Gadelha \textit{et al.}~\cite{gadelha2020label} decompose 3D objects into convex components and construct positive pairs among the same components and negative pairs among different components for contrastive learning.
Du \textit{et al.}~\cite{du2021self} introduced a hard negative sampling strategy into the contrastive learning between instances and local parts.
Besides, Rao \textit{et al.}~\cite{rao2020global} unified contrastive learning, normal estimation, and self-reconstruction into the same framework and formulated a multi-task learning method.

\begin{figure}[t]
    \centering
    \includegraphics[width=0.5\textwidth]{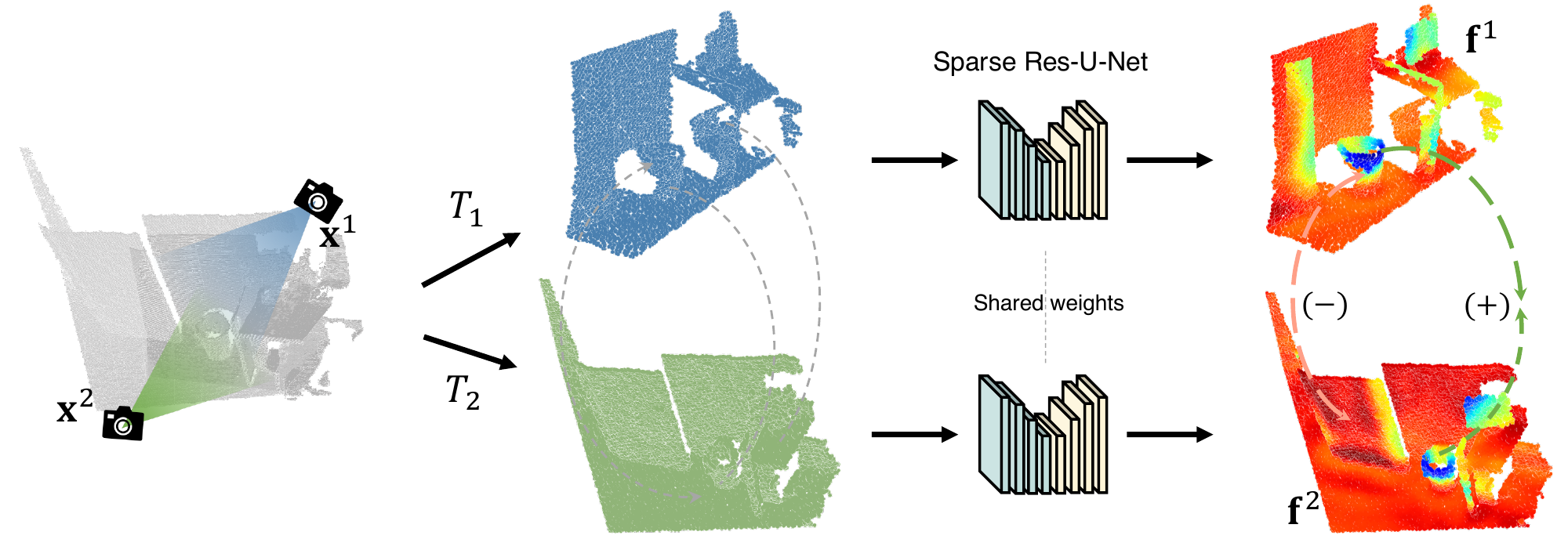}
    \caption{The pipeline of PointContrast \cite{xie2020pointcontrast}: Two scans $x^1$ and $x^2$ of the same scene captured from two different viewpoints are transformed by $T_1$ and $T_2$ for data augmentation. The correspondence mapping between the two views is computed to minimize the distance for matched point features and maximize the distance for unmatched point features for contrastive learning. The graph is extracted from \cite{xie2020pointcontrast} with authors’ permission.}
    \label{fig.PointContrast}
\end{figure}

Recently, Xie \textit{et. al} proposed PointContrast~\cite{xie2020pointcontrast}, a contrastive learning framework that learns representations of \textit{scene} point clouds as illustrated in Fig. \ref{fig.PointContrast}. The work shows, for the first time, that network weights pre-trained on 3D scene partial frames can lead to performance boosts when fine-tuned on multiple 3D high-level tasks including object classification, semantic segmentation, and object detection.
Firstly, dense correspondences are extracted between two aligned views of ScanNet~\cite{dai2017scannet} to build point pairs and point-level contrastive learning is then conducted with a unified backbone (SR-UNet). Finally, the learned model was transferred to multiple downstream 3D tasks including classification, semantic segmentation, and object detection with consistent performance gains.

Since PointContrast brought new insights that the unsupervised representation learned from scene-level point clouds can generalize across domains and boost high-level scene understanding tasks, several unsupervised pre-training works are proposed for scene-level 3D tasks. Considering that PointContrast focuses on point-level alignment without capturing spatial configurations and contexts in scenes, Hou \textit{et al.}~\cite{hou2021exploring} integrated spatial contexts into the pre-training objective by partitioning the space into spatially inhomogeneous cells for correspondence matching. Hou \textit{et al.}~\cite{Hou_2021_ICCV} built a multi-modal contrastive learning framework that models 2D multi-view correspondences as well as 2D-3D correspondences with geometry-to-image alignment. While the aforementioned works~\cite{xie2020pointcontrast,hou2021exploring,Hou_2021_ICCV} require 3D data captured from multiple camera views, Zhang \textit{et al.}~\cite{Zhang_2021_ICCV} proposed DepthContrast that can work with single-view data. Instead of using real point clouds as previous methods, Rao \textit{et al.}~\cite{rao2021randomrooms} generated synthetic scenes and objects from ShapeNet~\cite{chang2015shapenet}for network pre-training.

Another unsupervised approach to learn context similarity is \textit{clustering}. In this approach, samples are first grouped into clusters by clustering algorithms such as K-Means~\cite{hartigan1979algorithm} and each sample is assigned a cluster ID as pseudo-label. Then networks are trained in a supervised manner to learn semantic structures of data distribution. The learned parameters are used for model initialization for fine-tuning various downstream tasks. A typical example is DeepClustering~\cite{caron2018deep} which is the first unsupervised clustering method for 2D visual representation learning. However, no prior studies adopted a purely clustering strategy for URL of point clouds. Instead, hybrid approaches are proposed by integrating clustering with other unsupervised learning approaches (e.g., self-reconstruction~\cite{hassani2019unsupervised} or contrastive learning~\cite{zhang2019unsupervised}) for learning more robust representations.

\subsubsection{Learning with spatial context structure}

Point clouds with spatial coordinates provides accurate geometric description of 3D shapes of objects and scenes. The rich spatial contexts in point clouds can be exploited in pre-text tasks for URL. For example, networks can be trained to sort out the relation of different object parts. Likewise, the learned parameters can be used for model initialization for downstream tasks. Since no human annotations are required in training, the key is to design effective pre-text tasks to exploit spatial contexts as URL objectives.

\begin{figure}[t]
    \centering
    \includegraphics[width=0.45\textwidth]{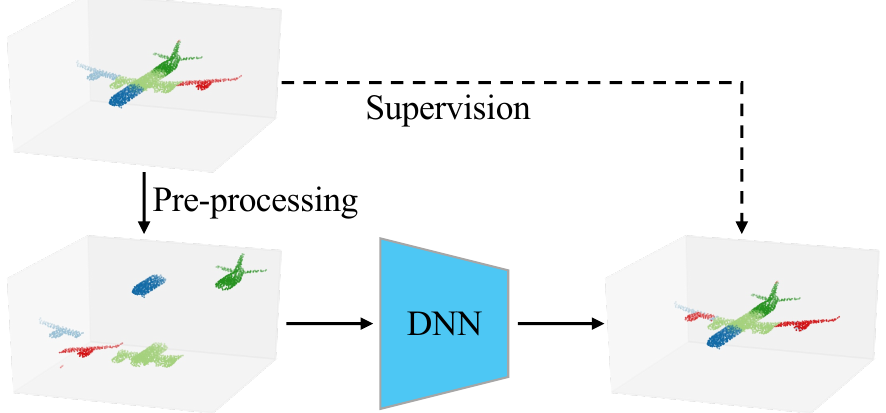}
    \caption{The pipeline of 3DJigsaw~\cite{sauder2019self}: An object is split into voxels where each point is assigned with a voxel label. The split voxels are randomly rearranged via pre-processing, and a deep neural network is trained to predict the voxel label for each point. The graph is reproduced based on \cite{sauder2019self}.}
    \label{fig.spatial context structure}
\end{figure}

The method Jigsaw3D~\cite{sauder2019self} proposed by Sauder \textit{et al.} is one of the pioneer works that use spatial context for URL of point clouds. As illustrated in Fig. \ref{fig.spatial context structure}, objects are first split into voxels where each point is assigned a voxel label. The network is then fed with randomly rearranged point clouds and optimized by predicting correct voxel label for each point. During the training, the network aims to extract spatial relations and geometric information from point clouds. In their following work~\cite{sauder2019context}, another pre-text task was designed to predict one of ten spatial relationships of two local parts from the same object. Inspired by the 2D method that predicts image rotations~\cite{gidaris2018unsupervised}, Poursaeed \textit{et al.}~\cite{poursaeed2020self} proposed to learn representations by predicting rotation angles of 3D objects. Thabet \textit{et al.}~\cite{thabet2020self} designed a pre-text task that predicts the next point in a point sequence defined by Morton-order Space Filling Curve. Chen \textit{et al.}~\cite{chen2021shape} proposed to learn the spatial context of objects by distinguishing the distorted parts of a shape from the correct ones. Sun \textit{et al.}~\cite{sun2021point} introduced a mix-and-disentangle task to exploit spatial context cues.

\subsubsection{Learning with temporal context structure}

Point cloud sequence is a common type of point cloud data that consists of consecutive point cloud frames. 
For example, there are indoor point cloud sequences transformed from RGB-D video frames~ \cite{dai2017scannet} and LiDAR sequential data \cite{geiger2013vision,behley2019semantickitti,xiao2022transfer} 
with continuous point cloud scans with each scan collected by one sweep of LiDAR sensors. Point cloud sequences contain rich temporal information that can be extracted by designing pre-text tasks and used as supervision signals to train DNNs. The learned representations can be transferred to downstream tasks.

\begin{figure}[t]
    \centering
    \includegraphics[width=0.5\textwidth]{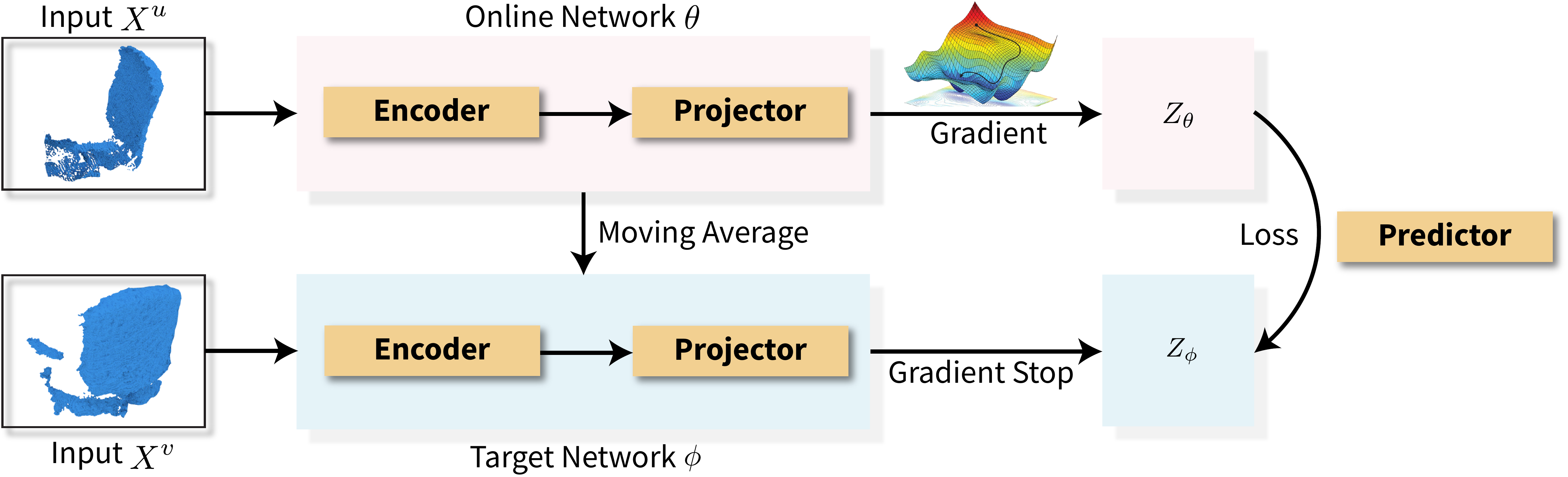}
    \caption{The pipeline of STRL~\cite{huang2021spatio}: An \textit{Online Network} learns spatial and temporal structures from two neighbouring point cloud frames $X^u$ and $X^v$. The figure is adopted from \cite{huang2021spatio} with authors' permission.}
    \label{fig.STRL}
\end{figure}

Recently, Huang \textit{et al.}~\cite{huang2021spatio} proposed a Spatio-Temporal Representation Learning (STRL) framework as illustrated in Fig. \ref{fig.STRL}. STRL extends BYOL~\cite{grill2020bootstrap} from 2D vision to 3D vision and extracts spatial and temporal representation from point clouds. It treats two neighboring point cloud frames as positive pairs and minimizes the mean squared error between the learned feature representations of sample pairs. Chen \textit{et al.}~\cite{chen20224dcontrast} exploit synthetic 3D shapes moving in static 3D environments to create dynamic scenarios and sample pairs in the temporal order. They conduct contrastive learning to learn 3D representations with dynamic understanding.

Unsupervised learning with temporal context structures has proved its effectiveness in both 2D computer vision tasks~\cite{feichtenhofer2021large,song2021spatio,hu2021contrast,kuang2021video} and 3D computer vision tasks~\cite{huang2021spatio,chen20224dcontrast}. As discussed in Section~\ref{Sec.Future}, this direction is very promising but more research is needed for better harvesting the temporal contextual information.

\subsection{Multiple modal-based methods}\label{sec.multi-modal}

Different modalities such as images \cite{geiger2013vision} and natural language descriptions \cite{chen2020scanrefer} can provide additional information for point-cloud data. 
Modeling relationships across modalities can be designed as pre-text tasks for URL which helps networks to learn more robust and comprehensive representations. Likewise, the learned parameters can be used as initialization weights for various downstream tasks.

\begin{figure}[t]
    \centering
    \includegraphics[width=0.5\textwidth]{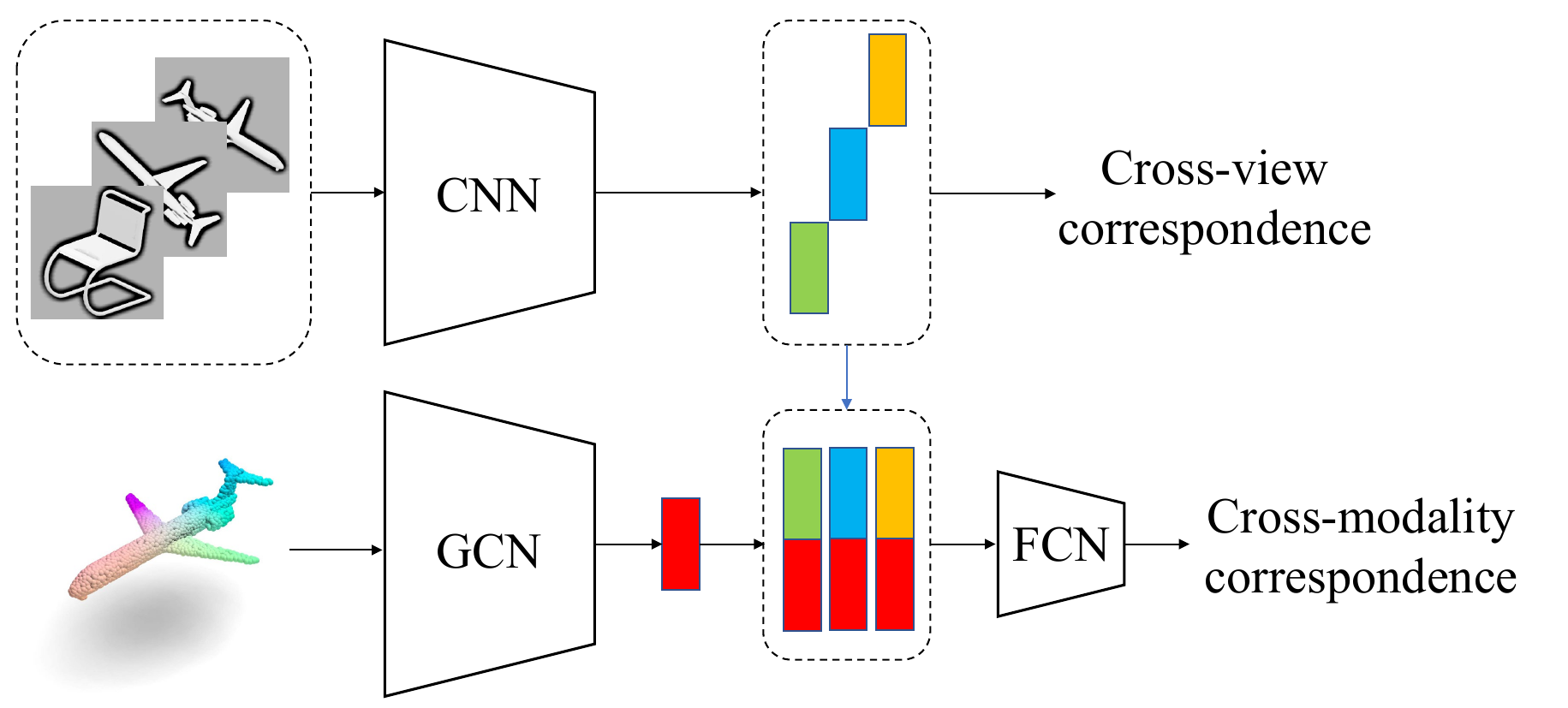}
    \caption{The pipeline CMCV~\cite{jing2021self}: CMCV employs a 2D CNN to extract 2D features from rendered views of 3D objects and a 3D GCN to extract 3D features from point clouds. The two types of features are concatenated by a two-layer fully connected network (FCN) to predict cross-modality correspondences. The graph is reproduced based on \cite{jing2021self}.}
    \label{fig.CMCV}
\end{figure}

Several recent work~\cite{jing2021self,afham2022crosspoint} exploits the correspondences across 3D point cloud objects and 2D images for URL. For example, Jing \textit{et al.}~\cite{jing2021self} render 3D objects with different camera views into 2D images for learning from multi-modality data. As Fig. \ref{fig.CMCV} shows, they employ a 2D CNN and a 3D GCN to extract image features and point cloud features, respectively, and then conduct contrastive learning on intra-modal correspondences and cross-modal correspondences. Their study shows that both pre-trained 2D CNN and 3D GCN achieved better classification as compared with random initialization. 
Differently, Wang et al.~\cite{wang2022p2p} project point clouds into colored images and then feed them into an \textit{image pre-trained} model with frozen weights to extract representative features for downstream tasks. 
However, how to learn unsupervised point cloud representations with other modalities such as text descriptions and audio data remains an under-explored field. We expect more studies in this promising research direction.

\subsection{Local descriptor-based methods}
The aforementioned methods aim to learn semantic structures of point clouds for high-level understanding, while the local descriptor-based methods focus on learning representations for low-level tasks. For example, Deng \textit{et al.}~\cite{deng2018ppf} introduced PPF-FoldNet that extracts rotation-invariant 3D local descriptors for 3D matching~\cite{zeng20173dmatch}. Several works~\cite{zeng2021corrnet3d,lang2021dpc} exploit non-rigid shape correspondence extraction as pre-text tasks for URL of point clouds, aiming to find the point-to-point correspondence of two deformable 3D shapes. Jiang \textit{et al.}~\cite{jiang2021sampling} explore unsupervised 3D registration for finding the optimal rigid transformation that can align the source point cloud to the target precisely. 

The performances of existing local descriptor-based methods are mainly evaluated on low-level tasks. However, how to adapt the learned feature representations toward other high-level tasks is rarely discussed. We expect more related research in the future.

\subsection{Pros and Cons}

\textbf{Generation-based methods} have been extensively studied in 3D URL, thanks to their ability to recover the original data distribution without assuming any downstream tasks. However, most existing research focuses on object-level point clouds, characterized by limited point numbers and data variability, restricting their applicability to object classification and part segmentation tasks. Additionally, these methods demonstrate limited effectiveness in scene-level tasks, such as 3D object detection and semantic segmentation, due to the difficulty of generating scene-level point clouds with complex distribution, rich noises and sparsity variation, and various occlusions.
Nonetheless, generation-based methods achieve very impressive progress in 2D images~\cite{he2022masked} recently, demonstrating their great potential for handling 3D point-cloud data. More efforts are expected in scene-level tasks as well as various downstream applications.

\noindent \textbf{Context-based methods} have recently become a prevalent approach in scene-level tasks, such as 3D semantic segmentation, 3D instance segmentation, and 3D object detection, thanks to their ability in addressing complex real-world data. However, they are still facing several challenges. The first is hard-example mining which is crucial to effective contrastive learning. 
Beyond that, designing effective self-supervision is also challenging for context-based methods, especially while considering generalization across various tasks and applications.

\noindent \textbf{Multiple modal-based methods} allow leveraging additional data modalities for enriching the distribution of point clouds. Pair-wise correspondences between point clouds and other data modalities also offer additional supervision, thereby enhancing the learned unsupervised point cloud representations. 
However, multi-modality methods are still facing several challenges. For example, acquiring large-scale pair-wise data is often a non-trivial task, and so does the design of effective cross-domain tasks. In addition, how to learn an effective homogeneous representation space across multiple modalities remains a very open research problem.

\noindent \textbf{Local descriptor-based methods} offer distinct advantages in capturing detailed spatial cues and exploiting low-level position information. However, these methods are limited in their ability of transferring learned representations to high-level recognition models, which restricts their application scope in more complex and abstract recognition tasks.

\section{Benchmark performances}\label{sec.Benchmarks}

\begin{table*}[!ht]
    \setlength\tabcolsep{10pt}
    \centering
    \caption{Comparing linear shape classification on ModelNet10 and ModelNet40 \cite{wu2015modelnet}: Linear SVM classifiers are trained with representations learned by different unsupervised methods. Accuracy highlighted by \textsuperscript{*} was obtained by pre-training with multi-modal data. [T] denotes models with modified Transformers. [ST] denotes models with standard Transformers.}
    \begin{tabular}{|l|c|c|c|c|c|c|c|}
        \hline
        Method & Year & Pre-text task & Backbone & Pre-train dataset & ModelNet10 & ModelNet40 \\
        \hline
        \multirow{6}{4em}{Supervised learning} & 2017 & \multirow{6}{2em}{N.A.} & PointNet~\cite{qi2017pointnet} & \multirow{4}{2em}{N.A.} & - & 89.2\\
        & 2017 &  & PointNet++~\cite{qi2017pointnet++} &  & - & 90.7\\
        & 2019 &  & DGCNN~\cite{wang2019dynamic} &  & - & 93.5\\
        & 2019 &  & RSCNN~\cite{liu2019rscnn} &  & - & 93.6 \\
        & 2021 & & [T]PointTransformer~\cite{zhao2021point} & & - & 93.7 \\
        & 2022 & & [ST]Transformer~\cite{yu2022point} & & - & 91.4 \\
        \hline
        SPH~\cite{kazhdan2003rotation} & 2003 & Generation & - & ShapeNet & 79.8 & 68.2\\
        LFD~\cite{chen2003visual} & 2003 & Generation & - & ShapeNet & 79.9 & 75.5\\
        TL-Net~\cite{girdhar2016tl_net} & 2016 & Generation & - & ShapeNet & - & 74.4\\
        VConv-DAE~\cite{sharma2016vconv-dae} & 2016 & Generation & - & ShapeNet & 80.5 & 75.5 \\
        3D-GAN~\cite{wu20163d-gan} & 2016 & Generation & - & ShapeNet & 91.0 & 83.3 \\
        3D DescriptorNet~\cite{xie2018learning} & 2018 & Generation & - & ShapeNet & - & 92.4 \\
        FoldingNet~\cite{yang2018foldingnet} & 2018 & Generation & - & ModelNet40  & 91.9 & 84.4 \\
        FoldingNet~\cite{yang2018foldingnet} & 2018 & Generation & - & ShapeNet & 94.4 & 88.4 \\
        Latent-GAN~\cite{achlioptas2018learning} & 2018 & Generation & - & ModelNet40 & 92.2 & 87.3 \\
        Latent-GAN~\cite{achlioptas2018learning} & 2018 & Generation & - & ShapeNet & 95.3 & 85.7 \\
        MRTNet~\cite{gadelha2018multiresolution} & 2018 & Generation & - & ShapeNet & 86.4 & -\\
        VIP-GAN~\cite{han2019view}  & 2019 & Generation & - & ShapeNet & 94.1 & 92.0\\
        3DCapsuleNet~\cite{zhao20193d} & 2019 & Generation & - & ShapeNet & - & 88.9\\
        PC-GAN~\cite{li2018point} & 2019 & Generation & - & ModelNet40 & - & 87.8\\
        L2G-AE~\cite{liu2019l2g} & 2019 & Generation & - & ShapeNet & 95.4 & 90.6\\
        MAP-VAE~\cite{han2019multi}& 2019  & Generation & - & ShapeNet & 94.8 & 90.2\\
        PointFlow~\cite{yang2019pointflow} & 2019 & Generation & - & ShapeNet & 93.7 & 86.8 \\
        MultiTask~\cite{hassani2019unsupervised} & 2019 & Hybrid & - & ShapeNet & - & 89.1\\
        Jigsaw3D~\cite{sauder2019self} & 2019 & Context & PointNet & ShapeNet & 91.6 & 87.3 \\
        Jigsaw3D~\cite{sauder2019self} & 2019 & Context & DGCNN & ShapeNet & 94.5 & 90.6 \\
        ClusterNet~\cite{zhang2019unsupervised} & 2019 & Context & DGCNN & ShapeNet & 93.8 & 86.8 \\
        CloudContext~\cite{sauder2019context} & 2019 & Context & DGCNN & ShapeNet & 94.5 & 89.3 \\
        NeuralSampler~\cite{remelli2019neuralsampler} & 2019 & Generation & - & ShapeNet & 95.3 & 88.7\\
        PointGrow~\cite{sun2020pointgrow} & 2020 & Generation & - & ShapeNet & 85.8 & -\\
        Info3D~\cite{sanghi2020info3d} & 2020 & Context & PointNet & ShapeNet & - & 89.8\\
        Info3D~\cite{sanghi2020info3d} & 2020  & Context & DGCNN & ShapeNet & - & 91.6\\
        ACD~\cite{gadelha2020label} & 2020 & Context & PointNet++ & ShapeNet & - & 89.8\\
        PDL~\cite{shi2020unsupervised} & 2020 & Generation & - & ShapeNet & - & 84.7\\
        GLR~\cite{rao2020global} & 2020 & Hybrid & PointNet++ & ShapeNet & 94.8 & 92.2 \\
        GLR~\cite{rao2020global} & 2020 & Hybrid & RSCNN & ShapeNet & 94.6 &  92.2\\
        SA-Net-cls~\cite{wen2020point} & 2020 & Generation & - & ShapeNet & - & 90.6 \\
        GraphTER~\cite{gao2020graphter} & 2020 & Generation & - & ModelNet40 & - & 89.1\\
        Rotation3D~\cite{poursaeed2020self} & 2020 & Context & PointNet & ShapeNet & - & 88.6\\ Rotation3D~\cite{poursaeed2020self} & 2020 & Context & DGCNN & ShapeNet & - & 90.8\\
        MID~\cite{wang2020unsupervised} & 2020 & Context & HRNet & ShapeNet & - & 90.3\\
        GTIF~\cite{chen2020deep} & 2020 & Generation & HRNet & ShapeNet & 95.9 & 89.6\\
        HNS~\cite{du2021self} & 2021 & Context & DGCNN & ShapeNet & - & 89.6\\
        ParAE~\cite{eckart2021self} & 2021 & Generation & PointNet & ShapeNet & - & 90.3 \\
        ParAE~\cite{eckart2021self} & 2021 & Generation & DGCNN & ShapeNet & - & 91.6 \\
        CMCV~\cite{jing2021self} & 2021 & Multi-modal & DGCNN & ShapeNet & - & 89.8\textsuperscript{*} \\
        GSIR~\cite{chen2021unsupervised} & 2021 & Context & DGCNN & ModelNet40 & - & 90.4\\
        STRL~\cite{huang2021spatio} & 2021 & Context & PointNet & ShapeNet & - & 88.3 \\
        STRL~\cite{huang2021spatio} & 2021 & Context & DGCNN & ShapeNet & - & 90.9\\
        PSG-Net~\cite{yang2021progressive} & 2021 & Generation & PointNet++ & ShapeNet & - & 90.9\\
        SelfCorrection~\cite{chen2021shape} & 2021 & Hybrid & PointNet & ShapeNet & 93.3 & 89.9\\
        SelfCorrection~\cite{chen2021shape} & 2021 & Hybrid & RSCNN & ShapeNet & 95.0 & 92.4\\
        OcCo~\cite{Wang_2021_ICCV} & 2021 & Generation & [ST]Transformer & ShapeNet & - & 92.1 \\
        CrossPoint~\cite{afham2022crosspoint} & 2022 & Multi-modal & PointNet & ShapeNet & - & 89.1\textsuperscript{*} \\
        CrossPoint~\cite{afham2022crosspoint} & 2022 & Multi-modal & DGCNN & ShapeNet & - & 91.2\textsuperscript{*} \\
        Point-BERT~\cite{yu2022point} & 2022 & Generation & [ST]Transformer & ShapeNet & - & 93.2 \\
        Point-MAE~\cite{pang2022masked} & 2022 & Generation & [ST]Transformer & ShapeNet & - & 93.8 \\
        \hline
    \end{tabular}
    \label{tab.linear classification}
\end{table*}

We benchmark representative 3D URL methods with two widely adopted evaluation metrics. The benchmarking is performed over public point-cloud data, where all performances are extracted from the corresponding papers.

\subsection{Evaluation Criteria}\label{subsec.eval_criteria}
There are two metrics that have been widely adopted for evaluating the quality of the learned unsupervised point-cloud representations.

\noindent$\bullet$ \textbf{Linear classification} first applies a pre-trained unsupervised model to extract features from certain labelled data. It then trains a supervised linear classifier with the extracted features together with the corresponding labels, where the quality of the pre-learned unsupervised representations is evaluated by the performance of the trained linear classifier over test data. Hence, the linear classification can be viewed as a type of representation learning metric which provides \textit{cluster analysis} in an implicit way.

\noindent$\bullet$ \textbf{Fine-tuning} optimizes a pre-trained unsupervised model using labelled data from downstream tasks. It can assess the quality of the pre-learned unsupervised representations by evaluating the performance of the fine-tuned model over downstream test data, \textit{i.e.} how much performance gains could be obtained by unsupervised pre-training compared to the random initialization.

\begin{table}[b]
    \setlength\tabcolsep{3pt}
    \centering
    \caption{Comparisons of unsupervised pre-training performance over the object classification datasets ModelNet40 and OBJ-BG split in ScanObjecNN. Performance numbers are presented in the format of "A/B", with "A" indicating training classification models from scratch with random initialization and "B" indicating fine-tuning classification models that are initialized with unsupervised pre-trained models. Performance under ``A" may vary due to different implementations as reported in the corresponding papers.}
    \begin{tabular}{|l|l|l|l|}
    \hline
        Method & Backbone & ModelNet40 & ScanObjectNN \\
    \hline
        Jigsaw3D~\cite{sauder2019self} & PointNet~\cite{qi2017pointnet} & 89.2/89.6\scriptsize{(+0.4)} & 73.5/76.5\scriptsize{(+3.0)}\\
        Info3D~\cite{sanghi2020info3d} & PointNet~\cite{qi2017pointnet} & 89.2/90.2\scriptsize{(+1.0)} & -/- \\
        SelfCorrection~\cite{chen2021shape} & PointNet~\cite{qi2017pointnet} & 89.1/90.0\scriptsize{(+0.9)} & -/- \\
        OcCo~\cite{Wang_2021_ICCV} & PointNet~\cite{qi2017pointnet} &  89.2/90.1\scriptsize{(+0.9)} & 73.5/80.0\scriptsize{(+6.5)} \\
        ParAE~\cite{eckart2021self} & PointNet~\cite{qi2017pointnet} &  89.2/90.5\scriptsize{(+1.3)} & -/- \\
        \hline
        Jigsaw3D~\cite{sauder2019self} & PCN~\cite{yuan2018pcn} &  89.3/89.6\scriptsize{(+0.3)} & 78.3/78.2\scriptsize{(-0.1)} \\
        OcCo~\cite{Wang_2021_ICCV} & PCN~\cite{yuan2018pcn} &  89.3/90.3\scriptsize{(+1.0)} & 78.3/80.4\scriptsize{(+2.1)} \\
        \hline
        GLR~\cite{rao2020global} & RSCNN~\cite{liu2019rscnn} & 91.8/92.2\scriptsize{(+0.5)}  & -/- \\
        SelfCorrection~\cite{chen2021shape} & RSCNN~\cite{liu2019rscnn} & 91.7/93.0\scriptsize{(+1.3)}  & -/- \\
        \hline
        Jigsaw3D~\cite{sauder2019self} & DGCNN~\cite{wang2019dynamic} &  92.2/92.4\scriptsize{(+0.2)} & 82.4/82.7\scriptsize{(+0.3)} \\
        Info3D~\cite{sanghi2020info3d} & DGCNN~\cite{wang2019dynamic} &  93.5/93.0\scriptsize{(-0.5)}  & -/- \\
        OcCo~\cite{Wang_2021_ICCV} & DGCNN~\cite{wang2019dynamic} &  92.5/93.0\scriptsize{(+0.5)} & 82.4/83.9\scriptsize{(+1.6)} \\
        ParAE~\cite{eckart2021self} & DGCNN~\cite{wang2019dynamic} &  92.2/92.9\scriptsize{(+0.7)} & -/- \\
        STRL~\cite{huang2021spatio} & DGCNN~\cite{wang2019dynamic} & 92.2/93.1\scriptsize{(+0.9)} & -/- \\
    \hline
        OcCo~\cite{Wang_2021_ICCV} & Transformer~\cite{yu2022point} & 91.2/92.2\scriptsize{(+1.0)} & 79.9/84.9\scriptsize{(+5.0)}  \\
        Point-BERT~\cite{yu2022point} & Transformer~\cite{yu2022point} & 91.2/93.4\scriptsize{(+2.2)} & 79.9/87.4\scriptsize{(+7.5)}  \\
    \hline
    \end{tabular}
    \label{tab:finetune-cls}
\end{table}

Note URL can be evaluated with other quantitative metrics. For example, \textit{reconstruction error}~\cite{yang2018foldingnet} can tell how well the learned representations encode the raw point clouds. Different clustering metrics such as \textit{Normalized Mutual Information}~\cite{hassani2019unsupervised} could complement the linear-classification metric. However, these metrics are mostly task-specific, e.g., the reconstruction error may not evaluate the representation of scene-level point clouds well due to their inherent noise, occlusion, and sparsity. In fact, few generic metrics can directly and explicitly evaluate the quality of the learned 3D unsupervised representations despite its critical importance to 3D URL studies. More research along this direction is needed to advance this research field further.

Beyond quantitative metrics, unsupervised feature representations can be evaluated in a qualitative manner. For example, t-SNE (t-Distributed Stochastic Neighbor Embedding)~\cite{van2008visualizing} has been widely adopted to compress the dimension of the learned feature representations and visualize the compressed feature embeddings.

\subsection{Object-level tasks}
\subsubsection{Object classification}

Object classification is the most widely used task in evaluations since the majority of existing works learn point cloud representations on object-level point cloud datasets. 
As described in Section \ref{subsec.eval_criteria}, both two types of protocols are widely adopted including the linear classification protocol and the fine-tuning protocol.

\begin{table}[!b]
    \setlength\tabcolsep{1pt}
    \centering
    \caption{Comparison of 3D URL methods for shape part segmentation over ShapeNetPart~\cite{chang2015shapenet}. \textbf{"Unsup."} denotes linear classification of the learned unsupervised point features. \textbf{"Trans."} is presented in a format of "A/B", where "A" is obtained with segmentation models trained from scratch with random initialization, and "B" is obtained by fine-tuning segmentation models that are initialized with unsupervised pre-trained models. We also provide supervised performances (\textbf{"Sup."}) of different backbone models with random initialization (extracted from the original papers).}
    \begin{tabular}{|l|c|c|c|c|cccc}
        \hline
        URL Method
        & Type
        & Backbone
        & class mIoU & instance mIoU \\
        \hline
        \multirow{5}{1em}{N.A.} & Sup. & PointNet & 80.4 & 83.7\\
         & Sup. & PointNet++ & 81.9 & 85.1\\
         & Sup. & DGCNN & 82.3 & 85.1\\
         & Sup. & RSCNN & 84.0 & 86.2\\
         & Sup. & Transformer & 83.4 & 85.1 \\
        \hline
        Latent-GAN~\cite{achlioptas2018learning} & Unsup. & - & 57.0 & - \\
        MAP-VAE~\cite{han2019multi}& Unsup. & - & 68.0 & - \\
        CloudContext~\cite{sauder2019context} & Unsup. & DGCNN & - & 81.5 \\
        GraphTER~\cite{gao2020graphter} & Unsup. & - & 78.1 & 81.9\\
        MID~\cite{wang2020unsupervised} & Unsup. & HRNet & 83.4 & 84.6\\
        HNS~\cite{du2021self} & Unsup. & DGCNN & 79.9 & 82.3\\
        CMCV~\cite{jing2021self} & Unsup. & DGCNN & 74.7 & 80.8 \\
        \hline
        SO-Net~\cite{li2018so-net} & Trans. & SO-Net & -/- & 84.6/84.9\scriptsize{(+0.3)}\\
        Jigsaw3D~\cite{sauder2019self} & Trans. & DGCNN & 82.3/83.1\scriptsize{(+0.8)} & 85.1/85.3\scriptsize{(+0.2)} \\
        MID~\cite{wang2020unsupervised} & Trans. & HRNet &  84.6/85.2\scriptsize{(+0.6)} & 85.5/85.8\scriptsize{(+0.3)}\\
        CMCV~\cite{jing2021self} & Trans. & DGCNN & 77.6/79.1\scriptsize{(+1.5)} & 83.0/83.7\scriptsize{(+0.7)}\\
        OcCo~\cite{Wang_2021_ICCV} & Trans. & PointNet & 82.2/83.4\scriptsize{(+1.2)} & -/-\\
        OcCo~\cite{Wang_2021_ICCV} & Trans. & DGCNN & 84.4/85.0\scriptsize{(+0.6)}  & -/-\\
        OcCo~\cite{Wang_2021_ICCV} & Trans. & Transformer & 83.4/83.4\scriptsize{(+0.0)}  & 85.1/85.1\scriptsize{(+0.0)}\\
        Point-BERT~\cite{yu2022point} & Trans. & Transformer & 83.4/84.1\scriptsize{(+0.7)}  & 85.1/85.6\scriptsize{(+0.5)}\\
        \hline
    \end{tabular}
    \label{tab.shapenet partseg}
\end{table}

Table \ref{tab.linear classification} summarizes the performance of the linear classification by existing methods. Specifically, linear classifiers are trained with the representations learned by different unsupervised methods on the ShapeNet or ModelNet40 dataset, and the classification results over the testing set over ModelNet10 and ModelNet40 are reported. For comparison, we also list supervised learning performances of the same backbone models over the same datasets. It can be seen that the performances of unsupervised learning methods keep improving and some methods have even surpassed supervised learning methods, demonstrating the effectiveness and great potential of URL of point clouds.

\begin{table*}[t]
    \setlength\tabcolsep{7pt}
    \centering
    \caption{Semantic segmentation on S3DIS~\cite{armeni2016s3dis}: It compares supervised training with random weight initialization and fine-tuning with pre-trained weights learned from unsupervised pre-training tasks. It uses DGCNN as the segmentation model,  which is trained on different single Areas and tested on Area 5 (upper part) and Area 6 (lower part).}
    \begin{tabular}{|l|l|l|l|l|l|l|l|l|l|l|}
        \hline
        \multirow{2}{4em}{Method} & 
        \multicolumn{5}{|c|}{OA on area \textbf{5} with different train area} & \multicolumn{5}{|c|}{mIoU on area \textbf{5} with different train area} \\
        \cline{2-11}
        & Area1 & Area2 & Area3 & Area4 & Area6 & Area1 & Area2 & Area3 & Area4 & Area6\\
        \hline
        \textit{from scratch} & 82.9 & 81.2 & 82.8 & 82.8 & 83.1 & 43.6 & 34.6 & 39.9 & 39.4 & 43.9\\
        Jigsaw3D~\cite{sauder2019self} & 83.5\scriptsize{(+0.6)} & 81.2\scriptsize{(+0.0)} & 84.0\scriptsize{(+1.2)} & 82.9\scriptsize{(+0.1)} & 83.3\scriptsize{(+0.2)} & 44.7\scriptsize{(+1.1)} & 34.9\scriptsize{(+0.3)} & 42.4\scriptsize{(+2.5)} & 39.9\scriptsize{(+0.5)} & 43.9\scriptsize{(+0.0)}\\
        ParAE~\cite{eckart2021self} & 91.8\scriptsize{(+8.9)} & 82.3\scriptsize{(+1.1)} & 89.5\scriptsize{(+6.7)} & 88.2\scriptsize{(+5.4)} & 86.4\scriptsize{(+3.3)} & 53.5\scriptsize{(+9.9)} & 38.5\scriptsize{(+3.9)} & 48.4\scriptsize{(+8.5)} & 45.0\scriptsize{(+5.6)} & 49.2\scriptsize{(+5.3)}\\
        \hline
        \hline
        \multirow{2}{4em}{Method} & 
        \multicolumn{5}{|c|}{OA on area \textbf{6} with different train area} & \multicolumn{5}{|c|}{mIoU on area \textbf{6} with different train area} \\
        \cline{2-11}
        & Area1 & Area2 & Area3 & Area4 & Area5 & Area1 & Area2 & Area3 & Area4 & Area5\\
        \hline
        \textit{from scratch} & 84.6 & 70.6 & 77.7 & 73.6 & 76.9 & 57.9 & 38.9 & 49.5 & 38.5 & 48.6\\
        STRL~\cite{huang2021spatio} & 85.3\scriptsize{(+0.7)} & 72.4\scriptsize{(+1.8)} & 79.1\scriptsize{(+1.4)} & 73.8\scriptsize{(+0.2)} & 77.3\scriptsize{(+0.4)} & 59.2\scriptsize{(+1.3)} & 39.2\scriptsize{(+0.8)} & 51.9\scriptsize{(+2.4)} & 39.3\scriptsize{(+0.8)} & 49.5\scriptsize{(+0.9)}\\
        \hline
    \end{tabular}
    \label{tab.S3DIS_semseg1}
\end{table*}

Table \ref{tab:finetune-cls} lists fine-tuning performance on the ModelNet40 and ScanObjectNN datasets. We can see that classification models initialized with unsupervised pre-trained weights always achieve better classification performances as compared with random initialization, regardless of backbone architectures. On the other hand, the performance gaps are still limited, largely due to the limited size and diversity of the pre-training datasets (i.e., ShapeNet and ModelNet40) and the simplicity of existing backbone models.
In comparison, thanks to the much larger pre-training datasets ImageNet~\cite{deng2009imagenet} and the more powerful backbone network ResNet ~\cite{he2016deep}, the state-of-the-art methods for unsupervised pre-training of 2D images are able to achieve more significant performance gains in the classification task. As discussed in Section \ref{Sec.Future}, we expect more diverse datasets and more advanced and generous backbone models that can set stronger foundations for this field.

\subsubsection{Object part segmentation}

Table \ref{tab.shapenet partseg} presents the benchmarking of object part segmentation on the ShapeNetPart dataset \cite{chang2015shapenet} using the linear classification protocol (i.e., "Unsup." in Table \ref{tab.shapenet partseg}) and the fine-tuning protocol (i.e., "Trans." in Table \ref{tab.shapenet partseg}) as described in Section~\ref{subsec.eval_criteria}. As the table shows, the performance gaps between unsupervised and supervised learning (i.e., "Unsup." vs. "Sup.") are decreasing. In addition, unsupervised pre-training achieves better performance in most cases under the fine-tuning protocol (i.e., "Trans." vs. "Sup."), though the improvement is still limited.

\begin{table}[t]
    \setlength\tabcolsep{13pt}
    \centering
    \caption{Performances for semantic segmentation on S3DIS~\cite{armeni2016s3dis}. Upper part: Models are tested on Area5 (Fold\#1) and trained on the rest of the data. Lower part: Six-fold cross-validation over three runs.}
    \begin{tabular}{|l|c|c|c|}
    \hline
        Method & Backbone & mACC & mIoU \\
    \hline
        \textit{from scratch} & \multirow{3}{4em}{SR-UNet} & 75.5  & 68.2 \\ 
        PointConstrast~\cite{xie2020pointcontrast} & & 77.0 & 70.9 \\
        DepthContrast~\cite{Zhang_2021_ICCV} & & - & 70.6 \\
    \hline
    \hline
    Method & Backbone & OA & mIoU \\
    \hline
        \textit{from scratch} & \multirow{3}{4em}{PointNet} & 78.2 & 47.0 \\
        Jigsaw3D~\cite{sauder2019self} & & 80.1 & 52.6 \\
        OcCo~\cite{Wang_2021_ICCV} & & 82.0 & 54.9 \\
    \hline
        \textit{from scratch} & \multirow{3}{4em}{PCN} & 82.9 & 51.1\\
        Jigsaw3D~\cite{sauder2019self} & & 83.7 & 52.2 \\
        OcCo~\cite{Wang_2021_ICCV} & & 85.1 & 53.4 \\
    \hline
        \textit{from scratch} & \multirow{3}{4em}{DGCNN} & 83.7 & 54.9\\
       Jigsaw3D~\cite{sauder2019self} &  & 84.1 & 55.6\\
       OcCo~\cite{Wang_2021_ICCV} &  &  84.6 & 58.0\\
    \hline
    \end{tabular}
    \label{tab.S3DIS_semseg2}
\end{table}

\subsection{Scene-level tasks}

As discussed in Section~\ref{Sec.Context-based}, unsupervised pre-training in scene-level tasks has recently become prevalent due to its enormous potential in various applications. This comes with a series of 3D URL studies that investigate the effectiveness of pre-training over different scene-level point cloud datasets. We provide a comprehensive benchmarking of these methods with respect to different 3D tasks.

\begin{table}[t]
    \setlength\tabcolsep{6pt}
    \centering
    \caption{Comparison of pre-training effects by different unsupervised learning methods. The benchmarking is 3D object detection task over datasets SUN RGB-D~\cite{song2015sun} and ScanNet-V2~\cite{dai2017scannet}. ``@0.25`` and ``@0.5`` represent per-category results of average precision (AP) with IoU threshold 0.25~(mAP@0.25) and 0.5~(mAP@0.5), respectively.}
    \begin{tabular}{|l|c|cc|cc|}
    \hline
        \multirow{2}{4em}{Method} & \multirow{2}{4em}{Backbone} & \multicolumn{2}{|c|}{SUN RGB-D} & \multicolumn{2}{|c|}{ScanNet-V2}\\
        \cline{3-6}
        & & @0.5 & @0.25 & @0.5 & @0.25 \\
    \hline
        \textit{from scratch} & \multirow{3}{4em}{SR-UNet} & 31.7 & 55.6 & 35.4 & 56.7 \\ 
        PointConstrast~\cite{xie2020pointcontrast} & & 34.8 & 57.5 & 38.0 & 58.5\\
        PC-FractalDB~\cite{Yamada_2022_CVPR} & & 35.9 & 57.1 & 37.0 & 59.4 \\
    \hline
        \textit{from scratch} & \multirow{7}{4em}{VoteNet} & 32.9 & 57.7 & 33.5 & 58.6 \\
        STRL~\cite{huang2021spatio} & & - & 58.2 & - & - \\
        RandRooms~\cite{rao2021randomrooms} & & 35.4 & 59.2 & 36.2 & 61.3\\    DepthContrast~\cite{Zhang_2021_ICCV} & & - & - & - & 62.2\\
        CSC~\cite{hou2021exploring} & & 33.6 & - & - & -\\
        PointContrast~\cite{xie2020pointcontrast} & & 34.0 & - & 38.0 & -\\
        4DContrast~\cite{chen20224dcontrast} & & 34.4 & - & 39.3 & - \\
    \hline
        \textit{from scratch} & \multirow{6}{5em}{PointNet++} & - & 57.5 & - & 58.6 \\
        PointContrast~\cite{xie2020pointcontrast} & & - & 57.9 & - & 58.5\\      
        RandRooms~\cite{rao2021randomrooms} & & - & 59.2 & - & 61.3\\   
        DepthContrast~\cite{Zhang_2021_ICCV} & & - & 60.7 & - & -\\
        PC-FractalDB~\cite{Yamada_2022_CVPR} & & 33.9 & 59.4 & 38.3 & 61.9 \\
        DPCo~\cite{li2022closer} &  & 35.6 & 59.8 & 41.5 & 64.2\\
    \hline
        \textit{from scratch} & \multirow{2}{4em}{H3DNet} & 39.0 & 60.1 & 48.1 & 67.3\\
        RandRooms~\cite{rao2021randomrooms} & & 43.1 & 61.6 & 51.5 & 68.6\\   
    \hline
    \end{tabular}
    \label{tab:detection}
\end{table}

\begin{table}[h]
    \setlength\tabcolsep{6.5pt}
    \centering
    \caption{Object detection performance on dataset ONCE \cite{mao2021one}. The baseline is trained from scratch. Unsupervised learning methods are used for pre-training models.
    $U_{small}$, $U_{median}$, and $U_{large}$ represent small, medium, and large amounts of unlabelled data that are used for unsupervised learning, respectively.}
    \begin{tabular}{|l|c|c|c|c|}
    \hline
        Method & Vehicle & Pedestrian & Cyclist & mAP \\
    \hline
        Baseline~\cite{yan2018second} & 69.7 & 26.1 & 59.9 & 51.9\\
    \hline
    \multicolumn{5}{c}{$U_{small}$}\\
    \hline
    BYOL~\cite{grill2020bootstrap} & 67.6 & 17.2 & 53.4 & 46.1 \scriptsize{(-5.8)}\\
    PointContrast~\cite{xie2020pointcontrast} & 71.5 & 22.7 & 58.0 & 50.8 \scriptsize{(-0.1)} \\
    SwAV~\cite{caron2020unsupervised} & 72.3 & 25.1 & 60.7 & 52.7 \scriptsize{(+0.8)}\\
    DeepCluster~\cite{caron2018deep} & 72.1 & 27.6 & 50.3 & 53.3 \scriptsize{(+1.4)}\\
    \hline
    \multicolumn{5}{c}{$U_{median}$}\\
    \hline
    BYOL~\cite{grill2020bootstrap} & 69.7 & 27.3 & 57.2 & 51.4 \scriptsize{(-0.5)}\\
    PointContrast~\cite{xie2020pointcontrast} & 70.2 & 29.2 & 58.9 & 52.8 \scriptsize{(+0.9)}\\
    SwAV~\cite{caron2020unsupervised} & 72.1 & 28.0 & 60.2 & 53.4 \scriptsize{(+1.5)}\\
    DeepCluster~\cite{caron2018deep} & 72.1 & 30.1 & 60.5 & 54.2 \scriptsize{(+2.3)}\\
    \hline
    \multicolumn{5}{c}{$U_{large}$}\\
    \hline
    BYOL~\cite{grill2020bootstrap} & 72.2 & 23.6 & 60.5 & 52.1 \scriptsize{(+0.2)}\\
    PointContrast~\cite{xie2020pointcontrast} & 73.2 & 27.5& 58.3 & 53.0 \scriptsize{(+1.1)}\\
    SwAV~\cite{caron2020unsupervised} &72.0 & 30.6 & 60.3& 54.3 \scriptsize{(+2.4)}\\
    DeepCluster~\cite{caron2018deep} & 71.9 & 30.5 & 60.4 & 54.3 \scriptsize{(+2.4)}\\
    \hline
    \end{tabular}
    \label{tab:det_once}
\end{table}

\begin{table}[t]
    \setlength\tabcolsep{12pt}
    \caption{Performances of instance segmentation on datasets S3DIS~\cite{armeni2016s3dis} and ScanNet-V2~\cite{dai2017scannet}. It reports the mean of average precision (mAP) across all semantic classes with a 3D IoU threshold of 0.25.}
    \centering
    \begin{tabular}{|l|c|c|c|}
    \hline
    
        Method & Backbone & S3DIS & ScanNet\\
    \hline
        \textit{from scratch} & \multirow{4}{4em}{SR-UNet} & 59.3 & 53.4\\
        PointContrast~\cite{xie2020pointcontrast} & & 60.5 & 55.8\\
        CSC~\cite{hou2021exploring} & & 63.4 & 56.5 \\
        4DContrast~\cite{chen20224dcontrast} & & - & 57.6 \\
    \hline
    \end{tabular}
    \label{tab:InstSeg}
\end{table}

Tables \ref{tab.S3DIS_semseg1} and \ref{tab.S3DIS_semseg2} show the performances of semantic segmentation on the S3DIS \cite{armeni2016s3dis} dataset.
We summarized them separately since different fine-tuning setups have been used in prior works. In Table \ref{tab.S3DIS_semseg1}, the unsupervised pre-trained DGCNN is fine-tuned on every \textit{single area} of S3DIS and tested on either Area 5 (the upper part of table) or Area 6 (the lower part of the table). Table \ref{tab.S3DIS_semseg2} instead shows the performance of fine-tuning different segmentation networks with the \textit{whole} dataset by following the one-fold (in the upper part of the table) and six-fold cross-validation setups (in the lower part of the table), respectively.

We also summarize existing works that handle unsupervised pre-training for object detection. Tables \ref{tab:detection} and \ref{tab:det_once} show their performances over indoor datasets including SUN RGB-D \cite{song2015sun} and ScanNet-V2 \cite{dai2017scannet} as well as outdoor LiDAR dataset ONCE \cite{mao2021one}, respectively. In addition, several works investigated unsupervised pre-training for instance segmentation. We summarize their performance over S3DIS \cite{armeni2016s3dis} and ScanNet-V2 \cite{dai2017scannet} in Table \ref{tab:InstSeg}.

It is inspiring to see that unsupervised learning representation can generalize across domains and boost performances over multiple  high-level 3D tasks as compared with training from scratch. These experiments demonstrate the huge potential of URL of point clouds in saving expensive human annotations. However, the improvements are still limited and we expect more research in this area.

\section{Future direction}\label{Sec.Future}

URL of point clouds has achieved significant progress during the last decade. We share several potential future research directions of this research field in this section.

\textbf{Unified 3D backbones are needed:} One major reason of the great success of deep learning in 2D computer vision is the standardization of CNN architectures with VGG~\cite{Simonyan15}, ResNet~\cite{he2016deep}, etc.
For example, the unified backbone structures greatly facilitate knowledge transfer across different datasets and tasks. For 3D point clouds, similar development is far under-explored, despite a variety of 3D deep architectures that have been recently reported. This can be observed from the URL methods in tables in Section \ref{sec.Benchmarks} most of which adopted very different backbone models. This impedes the development of 3D point cloud networks in scalable design and efficient deployment in various new tasks. Designing certain universal backbones that can be as ubiquitous as ResNet in 2D computer vision is crucial for the advance of 3D point cloud networks including unsupervised point cloud representation learning.

\textbf{Larger datasets are needed}: As described in Section~\ref{sec.PC_dataset}, most existing URL datasets were originally collected for the task of supervised learning. Since point cloud annotation is laborious and time-consuming, these datasets are severely constrained in data size and data diversity and are not suitable for URL with point clouds which usually requires large amounts of point clouds of good size and diversity. This issue well explains the trivial improvements by URL in tables in Section~\ref{sec.Benchmarks}. Hence, it is urgent to collect large-scale and high-quality unlabelled point cloud datasets of sufficient diversity in terms of object-level and scene-level point clouds, indoor and outdoor point clouds, etc.

\textbf{Unsupervised pre-training for scene-level tasks}: As described in Section~\ref{Sec.Context-based}, most earlier research focuses on object-level point cloud processing though several pioneer studies~\cite{xie2020pointcontrast,hou2021exploring,Hou_2021_ICCV, rao2021randomrooms, huang2021spatio} explored how to pre-train DNNs on scene-level point clouds for improving various scene-level downstream tasks such as object detection and instance segmentation. Prior studies show that the learned unsupervised representations can effectively generalize across domains and tasks. Hence, URL of scene-level point clouds deserves more attention as a new direction due to its great potential in a variety of applications. On the other hand, the research along this line remains at a nascent stage, largely due to the constraints in network architectures and datasets. We foresee that more related research will be conducted in the near future.

\textbf{Learning representations from multi-modal data:} 3D sensors are often equipped with other sensors that can capture additional and complementary information to point clouds. For example, depth cameras are often equipped with optical sensors for capturing better appearance information. LiDAR sensors, optical sensors, GPU, and IMU are often installed together as a sensor suite to capture complementary information and provide certain redundancy in autonomous vehicles and mobile robot navigation. Unsupervised learning from such multi-modal data has attracted increasing attention in recent years. For example, learning correspondences among multi-modal data has been explored as pre-text tasks for unsupervised learning as described in Section~\ref{sec.multi-modal}. However, the study along this line of research remains under-investigated and we expect more related research point clouds, RGB images, depth maps, etc.

\textbf{Learning Spatio-temporal representations:} 3D sensors that support capturing sequential point clouds are becoming increasingly popular nowadays. Rich temporal information from point cloud streams can be extracted as useful supervision signals for unsupervised learning while most of the existing works still focus on static point clouds. We expect that more effective pretext tasks will be designed that can effectively learn spatio-temporal representations from unlabelled sequential point cloud frames.

\section{Conclusion}
Unsupervised representation learning aims to learn effective representations from unannotated data, which has demonstrated impressive progress in the research with point cloud data. This paper presents a contemporary survey of unsupervised representation learning of point clouds. It first introduces the widely adopted datasets and deep network architectures. A comprehensive taxonomy and detailed review of methods are then presented. Following that, representative methods are discussed and benchmarked over multiple 3D point cloud tasks. Finally, we share our humble opinions about several potential future research directions. We hope that this work can lay a strong and sound foundation for future research in unsupervised representation learning from point cloud data.

\ifCLASSOPTIONcompsoc
  \section*{Acknowledgments}
\else
  \section*{Acknowledgment}
\fi

This project is funded in part by the Ministry of Education Singapore, under the Tier-1 scheme with project number RG18/22. It is also supported in part under the RIE2020 Industry Alignment Fund – Industry Collaboration Projects (IAF-ICP) Funding Initiative, as well as cash and in-kind contributions from Singapore Telecommunications Limited (Singtel), through Singtel Cognitive and Artificial Intelligence Lab for Enterprises (SCALE@NTU).

\ifCLASSOPTIONcaptionsoff
  \newpage
\fi



%

{\small
\bibliographystyle{IEEEtran}
\bibliography{ref_minor}
}

%
\vspace{-1cm}

\begin{IEEEbiography}[{\includegraphics[width=1in,height=1.25in,clip,keepaspectratio]{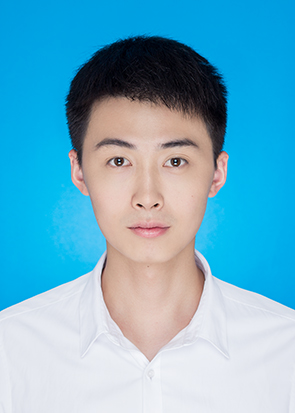}}]{Aoran Xiao}
received his B.Sc. and M.Sc. degree from Wuhan University, China in 2016 and 2019, respectively. He is currently pursuing the Ph.D. degree with the school of computer science and engineering
at Nanyang Technological University, Singapore. His research interests lie in point cloud processing, computer vision, and remote sensing.
\end{IEEEbiography}

\vspace{-2cm}

\begin{IEEEbiography}[{\includegraphics[width=1in,height=1.25in,clip,keepaspectratio]{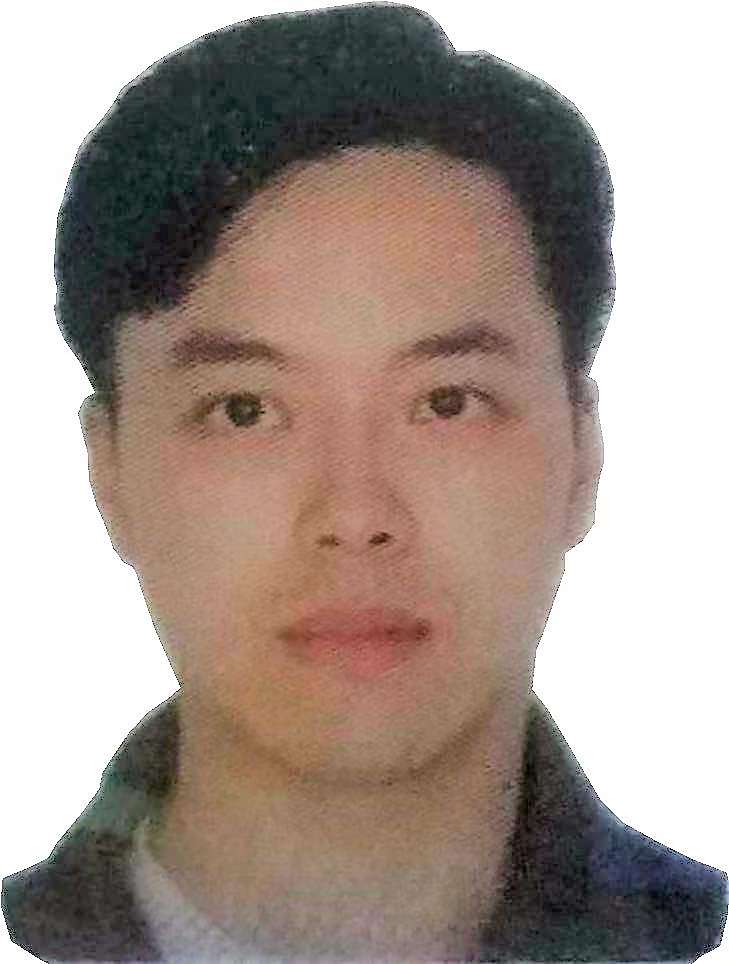}}]{Jiaxing Huang} received his B.Eng. and M.Sc. in EEE from the University of Glasgow, UK, and the Nanyang Technological University (NTU), Singapore, respectively. He is currently a Research Associate and Ph.D. student with School of Computer Science and Engineering, NTU, Singapore. His research interests include computer vision and machine learning.
\end{IEEEbiography}

\vspace{-2cm}

\begin{IEEEbiography}[{\includegraphics[width=1in,height=1.25in,clip,keepaspectratio]{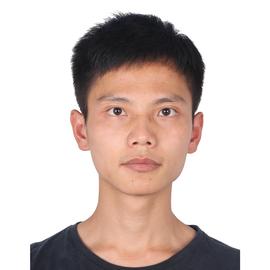}}]{Dayan Guan} is currently a Research Scientist at Mohamed bin Zayed University of Artificial Intelligence, United Arab Emirates. Before that, he had been a Research Fellow at Nanyang Technological University from Nov 2019 to Mar 2022. In Sep 2019, he received his Ph.D. from Zhejiang University, China. His research interests include computer vision, pattern recognition and deep learning.
\end{IEEEbiography}

\vspace{-2cm}

\begin{IEEEbiography}[{\includegraphics[width=1in,height=1.25in,clip,keepaspectratio]{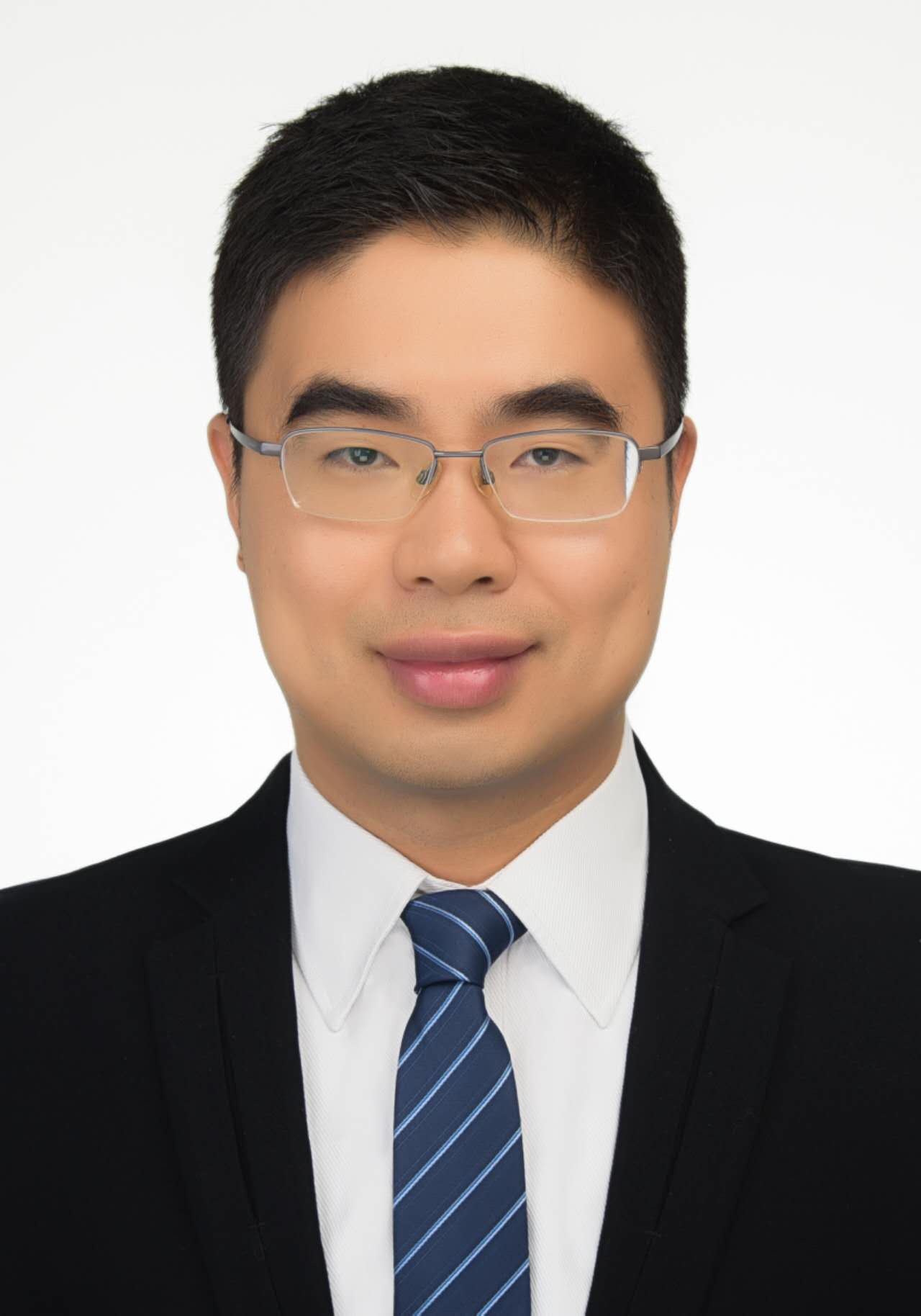}}]{Xiaoqin Zhang} is a senior member of the IEEE. He received the B.Sc. degree in electronic information science and technology from Central South University, China, in 2005, and the Ph.D. degree in pattern recognition and intelligent system from the National Laboratory of Pattern Recognition, Institute of Automation, Chinese Academy of Sciences, China, in 2010. He is currently a Professor with Wenzhou University, China. He has published more than 100 papers in international and national journals, and international conferences, including IEEE T-PAMI, IJCV, IEEE T-IP, IEEE T-NNLS, IEEE T-C, ICCV, CVPR, NIPS, IJCAI, AAAI, and among others. His research interests include in pattern recognition, computer vision, and machine learning.
\end{IEEEbiography}

\vspace{-2cm}

\begin{IEEEbiography}[{\includegraphics[width=1in,height=1.25in,clip,keepaspectratio]{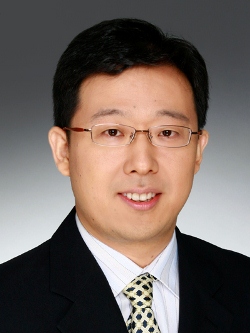}}]{Shijian Lu} is an Associate Professor with the School of Computer Science and Engineering at the Nanyang Technological University, Singapore. He received his PhD in electrical and computer engineering from the National University of Singapore. His major research interests include image and video analytics, visual intelligence, and machine learning.
\end{IEEEbiography}

\vspace{-2cm}

\begin{IEEEbiography}[{\includegraphics[width=1in,height=1.25in,clip,keepaspectratio]{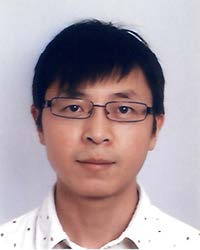}}]{Ling Shao} is the Chief Scientist of Terminus Group and the President of Terminus International. He was the founding CEO and Chief Scientist of the Inception Institute of Artificial Intelligence, Abu Dhabi, UAE. His research interests include computer vision, deep learning, medical imaging and vision and language. He is a fellow of the IEEE, the IAPR, the BCS and the IET.
\end{IEEEbiography}





\end{document}